%% file: main.tex
\documentclass[10pt,journal,compsoc]{IEEEtran}
\usepackage[utf8]{inputenc} 
\usepackage[T1]{fontenc}    
\usepackage{hyperref}       
\usepackage{url}            
\usepackage{booktabs}       
\usepackage{amsfonts}       
\usepackage{nicefrac}       
\usepackage{microtype}      
\usepackage[dvipsnames, svgnames, x11names]{xcolor}
\usepackage{graphicx}
\usepackage{amsmath}
\usepackage{subfigure}
\usepackage{tcolorbox}
\usepackage{multirow}
\usepackage{pifont}
\usepackage{enumitem}
\usepackage{amssymb}
\newcommand{\tabincell}[2]{\begin{tabular}{@{}#1@{}}#2\end{tabular}}
\usepackage{enumitem}
\usepackage[misc]{ifsym}

\ifCLASSOPTIONcompsoc
  \usepackage[nocompress]{cite}
\else
  \usepackage{cite}
\fi
\ifCLASSINFOpdf
\else
\fi

\def\eg{\emph{e.g.}}
\def\ie{\emph{i.e.}}
\def\etc{\emph{etc}}
\def\vs{\emph{v.s.}}
\def\wrt{\emph{w.r.t.}}

\newif\ifsubmit
\submitfalse

\begin{document}
%
\title{\emph{Robust}$\mathcal{ART}$: Benchmarking Robustness on Architecture Design and Training Techniques}
%
%
%
%

\author{%

  Shiyu Tang\textsuperscript{*}, Ruihao Gong\textsuperscript{*}\footnote[1]{}, Yan Wang\textsuperscript{*}\footnote[1]{}, Aishan Liu\textsuperscript{* \Letter}, Jiakai Wang, Xinyun Chen, \\
  Fengwei Yu, Xianglong Liu\textsuperscript{\Letter}, Dawn Song, Alan Yuille, Philip H.S. Torr, Dacheng Tao\textsuperscript{\Letter}\\
  

\thanks{S. Tang, A. Liu, J. Wang, and X. Liu are with the State Key Lab of Software Development Environment, Beihang University, Beijing 100191, China.}

\thanks{R. Gong, Y. Wang, F. Yu are with the SenseTime, Beijing 100191, China.}

\thanks{X. Chen, D. Song are with the Berkeley Artificial Intelligence Research Lab, University of California, Berkeley, Berkeley, CA 94701 USA.}

\thanks{A. Yuille is with the Department of Cognitive Science and Computer Science, Johns Hopkins University, 	Baltimore, MD 21218 USA.}

\thanks{PH. Torr is with the Department of Engineering Science, University of Oxford, Oxford, OX1 3PJ UK.}

\thanks{D. Tao is with the JD Explore Academy, Beijing 101111, China.}

\thanks{* indicates equal contributions. \Letter\,indicates corresponding author.}

}
%
%

\markboth{IEEE TRANSACTIONS ON PATTERN ANALYSIS AND MACHINE INTELLIGENCE}%
{Shell \MakeLowercase{\textit{et al.}}: Bare Demo of IEEEtran.cls for Computer Society Journals}
%



\IEEEtitleabstractindextext{%
\begin{abstract}
  Deep neural networks (DNNs) are vulnerable to adversarial noises, which motivates the benchmark of model robustness. Existing benchmarks mainly focus on evaluating defenses, but there are no comprehensive studies of how architecture design and training techniques affect robustness. Comprehensively benchmarking their relationships is beneficial for better understanding and developing robust DNNs. Thus, we propose \textbf{RobustART}, the first comprehensive \textbf{Robust}ness investigation benchmark on ImageNet regarding \textbf{AR}chitecture design (49 human-designed off-the-shelf architectures and 1200+ networks from neural architecture search) and \textbf{T}raining techniques (10+ techniques, \eg, data augmentation) towards diverse noises (adversarial, natural, and system noises). Extensive experiments substantiated several insights for the first time, \eg, (1) adversarial training is effective for the robustness against all noises types for Transformers and MLP-Mixers; (2) given comparable model sizes and aligned training settings, CNNs > Transformers > MLP-Mixers on robustness against natural and system noises; Transformers > MLP-Mixers > CNNs on adversarial robustness; (3) for some light-weight architectures, increasing model sizes or using extra data cannot improve robustness. Our benchmark \url{http://robust.art/} presents: (1) an open-source platform for comprehensive robustness evaluation; (2) a variety of pre-trained models to facilitate robustness evaluation; and (3) a new view to better understand the mechanism towards designing robust DNNs. We will continuously develop to this ecosystem for the community.
\end{abstract}

\begin{IEEEkeywords}
Model Robustness, Robustness Investigation Benchmark, Computer Vision.
\end{IEEEkeywords}}

\maketitle

\IEEEdisplaynontitleabstractindextext

%
\IEEEpeerreviewmaketitle

\input{intro}
\input{relatedworks}

\input{benchmark}

\input{exp}
\input{conclu}

\ifCLASSOPTIONcaptionsoff
  \newpage
\fi

{
\bibliography{reference}
\bibliographystyle{unsrt}
}


\newpage
\newpage
\clearpage

\appendices
\input{supp}

\end{document}

%% file: intro.tex
\IEEEraisesectionheading{\section{Introduction}}

\IEEEPARstart{D}{eep} neural networks (DNNs) have achieved remarkable performance across a wide range of applications \cite{Krizhevsky2012ImageNet,bahdanau2014neural,Hinton2012Deep}. However, DNNs are susceptible to \emph{adversarial examples} \cite{szegedy2013intriguing,goodfellow6572explaining}; i.e., adding deliberately crafted perturbations imperceptible to humans could easily lead DNNs to wrong predictions, threatening both digital and physical deep learning applications \cite{kurakin2016adversarial,eykholt2018robust,Liu2019Perceptual,Liu2020Biasbased, arnab2020robustness,bai2020adversarial}. Besides, some prior works show that DNNs are also vulnerable to natural noises \cite{hendrycks2018benchmarking,taori2020measuring,hendrycks2021natural,hendrycks2020many}. These phenomena demonstrate that deep learning systems are not inherently secure and robust.

Since DNNs have been integrated into various safety-critical scenarios (\eg, autonomous driving, face recognition)~\cite{kendall2019learning,grigorescu2020survey,parkhi2015deep,sun2015deepid3}, improving model robustness and further constructing robust deep learning systems in practice is now growing in importance. 
Benchmarking the robustness of deep learning models paves a critical path to better understanding and further improving model robustness \cite{carlini2019evaluating,Ling2019Deepsec,Ma2018DeepGauge,Dong2020Benchmarking}. Existing benchmarks focus on evaluating the performance of commonly used adversarial defense methods \cite{madry2017towards,zhang2019theoretically,Shafahi2019Free,liu2021ANP}, however, there are no comprehensive studies of how architecture design and general training techniques affect robustness. These factors reflect the inherent nature of model robustness, and slight differences may override the gains from the defenses \cite{yao2018hessian,kamath2020sgd}. Thus, a comprehensive benchmark and study of the influence of architecture design and training techniques on robustness are highly important for understanding and improving the robustness of deep learning models. 


In this work, we propose \textbf{RobustART}, the first comprehensive robustness benchmark on ImageNet regarding architecture design and training techniques towards diverse noise types. We systematically study over a thousand model architectures (\emph{49} prevalent human-designed off-the-shelf model architectures including CNNs~\cite{he2016deep,huang2017densely,tan2019efficientnet}, Transformers \cite{DBLP:conf/iclr/DosovitskiyB0WZ21,touvron2020training,liu2021swin}, and MLP-Mixers \cite{tolstikhin2021mlp}; \emph{1200+} architectures generated by Neural Architecture Search \cite{xie2017genetic,Liu_2018_ECCV,yu2020bignas,li2020neural,mei2019atomnas}) and \emph{10+} mainstream training techniques (\eg, data augmentation, adversarial training, weight averaging, label smoothing, optimizer choices). To thoroughly study robustness against noises from different sources, we evaluate diverse noise types including adversarial noises \cite{goodfellow6572explaining,madry2017towards}, natural noises (\eg, ImageNet-C \cite{hendrycks2018benchmarking}, -P \cite{hendrycks2018benchmarking}, -A \cite{hendrycks2021natural}, and -O \cite{hendrycks2021natural}), and system noises \cite{wang2021imagenets} (see Table \ref{tab:architecture-rank} for the top-5 robust architectures against different noises). Our large-scale experiments revealed several insights with respect to general model design for the first time: (1) for Transformers and MLP-Mixers, adversarial training is universally effective for improving the robustness against all types of noises (adversarial, natural, and system noises); (2) given comparable model sizes and aligned training settings, CNNs (\eg, ResNet) perform stronger on natural and system noises, while Transformers (\eg, ViT) are more robust on adversarial noises; and (3) for almost all model families, increasing model capacity improves model robustness; however, increasing model sizes or adding extra data cannot improve robustness for some lightweight architectures. 

Our contributions are as follows.
\begin{itemize}
\setlength{\parskip}{-0.015in}
\item{\textbf{Comprehensive benchmark.} We provide the first comprehensive robustness benchmark on ImageNet, regarding architecture design and general training techniques. We investigate more than one thousand model architectures and \emph{10+} training techniques on multiple noise types (adversarial, natural, and system noises).}

\item{\textbf{Open-source framework.} We open-source our whole framework, which consists of a \emph{model zoo} (100+ pre-trained models), an open-source \emph{toolkit}, and all \emph{datasets} (\ie, ImageNet with different noise types). The framework provides an open platform for the community, enabling comprehensive robustness evaluation.}

\item{\textbf{In-depth analyses.} Based on extensive experiments, we revealed and substantiated several insights that reflect the inherent relationship between robustness and architecture design with different training techniques.}
\end{itemize}

Our benchmark \url{http://robust.art/} provides an open-source platform and framework for comprehensive evaluations, better understanding of DNNs, and robust architectures design. Together with existing benchmarks on defenses, we could build a more comprehensive robustness benchmark and ecosystem involving more perspectives.

\begin{table*}[t]
\centering
\caption{Top-5 robust model architectures on ImageNet for different noises. For fair comparisons, we keep the aligned training settings for all architectures.}
\label{tab:architecture-rank}
\resizebox{1\linewidth}{!}{
\begin{tabular}{cc|ccccc}
\toprule
\multirow{2}{*}{\textbf{Noise Type}}               & \multirow{2}{*}{\textbf{Datasets / Attacks}} & \multicolumn{5}{c}{\textbf{Model Performance Ranking}}                                 \\
                                             &                                              & No.1             & No.2              & No.3          & No.4           & No.5           \\ \midrule
\textbf{Clean Images}                        & ImageNet                                     & ResNeXt-101       & WideResNet-101     & RegNetX-6400M & ResNet-152      & Swin-B      \\ \midrule
\multirow{2}{*}{\textbf{Adversarial Noises}} & AutoAttack-$\ell_{\infty}$       & ViTAE-S            & ViT-B/32         & DeiT-B            & ViT-B/16   & ViTAE-T  \\
                                             & Worst-Case Adv                            & ViT-B/32   & ViTAE-S      & DeiT-B            & ViT-B/16      & MLP-Mixer L/16  \\ \midrule
\multirow{4}{*}{\textbf{Natural Noises}}     & ImageNet-C                                   & ResNeXt-101       & WideResNet-101     & ResNet-152     & ResNet-101      & RepVGG-B3      \\
                                             & ImageNet-P                                   & ResNeXt-101 & WideResNet-101 & ResNet-152    & RepVGG-B3  & ResNet-101      \\
                                             & ImageNet-A                                   & Swin-B       &    ResNeXt-101      &     Swin-S & RegNetX-6400M  & ResNet-152  \\
                                             & ImageNet-O                                   & ResNeXt-101       & Swin-B            & Swin-T  & ViTAE-S     & Swin-S           \\ \midrule
\textbf{System Noises}                       & ImageNet-S                                   & EfficientNet-B4      & DenseNet-201         & ResNeXt-50    &   ResNeXt-101    & ResNet-152      \\ \bottomrule
\end{tabular}
}
\end{table*}

%% file: relatedworks.tex
\section{Background and Related Work}
In this section, we provide a brief overview of existing work on adversarial attacks and defenses, as well as robustness benchmark and evaluation.

\subsection{Adversarial attacks and defenses}
\label{sec:adv}
Adversarial examples are inputs intentionally designed to mislead DNNs \cite{szegedy2013intriguing,goodfellow6572explaining}. Given a DNN $f_{\Theta}$ and an input image $\mathbf{x} \in \mathbb{X}$ with the ground truth label $\mathbf{y} \in \mathbb{Y}$, an adversarial example $\mathbf{x}_{adv}$ satisfies

\begin{equation}
f_{\Theta}(\mathbf x_{adv}) \neq \mathbf y  \quad s.t. \quad \|\mathbf x-\mathbf x_{adv}\| \leq \epsilon,
\end{equation}

where $\|\cdot\|$ is a distance metric. Commonly, $\|\cdot\|$ is measured by the $\ell_{p}$-norm ($p\in$\{1,2,$\infty$\}).

In particular, adversarial attacks can be divided into two types: (1) white-box attacks, in which adversaries have the complete knowledge of the target model and can fully access it; and (2) black-box attacks, in which adversaries do not have full access to the target model and only have limited knowledge of it, \eg, can only obtain its prediction without knowing the architecture and weights. A plethora of work has been devoted to performing adversarial attacks \cite{goodfellow6572explaining,athalye2018obfuscated,Jonathan2018SPSA,croce2020autoattack,mopuri2018generalizable,wang2020hamiltonian,dong2021query}. Goodfellow et al. \cite{goodfellow6572explaining} first proposed the Fast Gradient Sign Method (FGSM) which leverages the gradient information of input image to craft adversarial examples efficiently. Madry et al. \cite{madry2017towards} proposed the Project Gradient Descent (PGD) attack that is similar with FGSM but iterates more steps to generate adversarial examples, after each step it projects noises to the $\epsilon$-ball around input images. Carlini \& Wagner (C\&W) attack \cite{carlini2017towards} is an optimization-based attack that aims to find adversarial perturbations to minimize the specific object function. It transforms a general constrained optimization problem into minimizing an empirically-chosen object function. Universal Adversarial Perturbation (UAP) \cite{moosavi2017universal} aims to find the input-agnostic noises that could cause model misclassification when added on input images from various classes. AutoAttack \cite{croce2020autoattack} is one of the state-of-the-art adversarial attacks that combine four different attacks to form a perameter-free and computationally affordable attack to test adversarial robustness. It includes two extensions of PGD attack (APGD-CE and APGD-DLR) and two existing adversarial attacks (FAB attack \cite{croce2020fab} and Square Attack \cite{ACFH2020square}) to further boost the strength of attack. 

On the other hand, various defense approaches have been proposed to improve model robustness against adversarial examples \cite{Papernot2015Distillation,xie2018mitigating,madry2017towards,liao2018defense,pmlr-v70-cisse17a,goodfellow6572explaining,liu2021ANP,mustafa2020deeply,theagarajan2021privacy}. Specifically, adversarial training minimizes the worst case loss within some perturbation region for classifiers, by augmenting the training set $\{x^{(i)}, y^{(i)}\}_{i=1...n}$ with adversarial examples. Defensive Distillation \cite{Papernot2015Distillation} uses the change of softmax temperature to obsfucate the gradient information of model outputs and thus prevents the white-box attack from using gradient to generate adversarial examples. Zhang et al. \cite{zhang2019theoretically} stated the trade-off between robustness and accuracy and proposed TRADES to defend adversarial attacks. TRADES introduces a regularization term to induce a shift of the decision boundaries away from the training data points. Kannan et al. proposed Adversarial Logit Pairing (ALP) \cite{kannan2018adversarial} which not only minimizes the model prediction error on adversarial examples but also tries to minimize the distance of model logits between clean examples and the corresponding adversarial examples.

\subsection{Robustness benchmark and evaluation}
A number of works have been proposed to evaluate the robustness of deep neural networks \cite{carlini2019evaluating,Ling2019Deepsec,xie2019denoising,xie2020smooth,Dong2020Benchmarking,pang2021bag,mahmood2021On,shao2021Vit,Paul2021Vision,Aldahdooh2021Reveal}. Su et al. \cite{su2018is} first investigated the adversarial robustness of 18 models on ImageNet. DEEPSEC \cite{Ling2019Deepsec} is a platform for adversarial robustness analysis, which incorporates 16 adversarial attacks, 13 adversarial defenses, and several utility metrics. Carlini et al. \cite{carlini2019evaluating} discussed the methodological foundations, reviewed commonly accepted best practices, and the suggested checklist for evaluating adversarial defenses. RealSafe \cite{Dong2020Benchmarking} is another benchmark for evaluating adversarial robustness on image classification tasks (including 15 attacks and 16 defenses). Pang et al. \cite{pang2021bag} conducted a thorough empirical study of the training tricks of representative adversarial training methods to improve the robustness on CIFAR-10. More recently, RobustBench \cite{croce2020robustbench} is developed to track and evaluate the state-of-the-art adversarial defenses on CIFAR-10 and CIFAR-100.

Compared to existing benchmarks, our benchmark has the following characteristics: (1) comprehensive evaluation of different model architectures under aligned training settings and general training schemes over different architectures, while prior works focus on training schemes specialized for improving the performance against limited noise types, \eg, defenses against adversarial examples or common corruptions; (2) all noise types are based on ImageNet, while prior benchmarks mainly focus on image datasets of much smaller sizes, \eg, CIFAR-10 and CIFAR-100; and (3) systematic study of diverse noise types, while prior benchmarks mainly focus on evaluating adversarial robustness.

%% file: benchmark.tex
\section{Robustness Benchmark on Architecture Design and Training Techniques}
\label{sec:big_benchmark}

Existing robustness benchmarks mainly evaluate adversarial defenses, and lack a comprehensive study of the effect of architecture design and general training techniques on robustness. These factors reflect the inherent nature of the model, and comprehensively benchmarking their effects on robustness lays the foundation for building robust DNNs. 
Therefore, we propose the first comprehensive robustness benchmark on ImageNet considering architecture design and training techniques.

\subsection{Factors affecting robustness}
Our main goal is to investigate the effects of two orthogonal factors on robustness, which are \textit{architecture design} and \textit{training techniques}. Therefore, we build a comprehensive repository containing: (1) models with different architectures, but trained with the same (aligned) techniques; and (2) models with the same architecture, but trained with different techniques. These fine-grained ablation studies enable a better understanding of how the core model design choices contribute to the robustness.

\begin{table*}[t]
\centering
\caption{Architecture repository.}
\label{tab:architecture}
\resizebox{1\linewidth}{!}{
\begin{tabular}{l l l}
\toprule
\textbf{Category} & \textbf{Arch. Family} & \textbf{Archs} \\ 
\midrule

\multirow{10}{*}{CNN} & ResNet \cite{he2016deep} & ResNet-18, ResNet-34, ResNet-50, ResNet-101, ResNet-152  \\
                     & ResNeXt \cite{xie2017aggregated} & ResNeXt-50-32x4d, ResNeXt-101-32x8d \\
                     & WideResNet \cite{zagoruyko2016wide} & WRN-50-2, WRN-101-2 \\
                     & DenseNet \cite{huang2017densely} & DenseNet-121, DenseNet-169, DenseNet-201 \\
                     & ShuffleNetV2 \cite{ma2018shufflenet} & ShuffleNetV2-x0.5, ShuffleNetV2-x1.0, ShuffleNetV2-x1.5, ShuffleNetV2-x2.0 \\
                     & MobileNetV2 \cite{sandler2018mobilenetv2} & MobileNetV2-x0.5, MobileNetV2-x0.75, MobileNetV2-x1.0, MobileNetV2-x1.4 \\
                     & RepVGG \cite{ding2021repvgg} & RepVGG-A0, RepVGG-B3 \\
                     & RegNetX \cite{radosavovic2020designing} & RegNetX-400M, RegNetX-800M, RegNetX-1600M, RegNetX-3200M, RegNetX-6400M \\
                     & EfficientNet \cite{tan2019efficientnet} & EfficientNet-B0, EfficientNet-B1, EfficientNet-B2, EfficientNet-B3, EfficientNet-B4 \\
                     & MobileNetV3 \cite{howard2019searching} & MobileNetV3-x0.35, MobileNetV3-x0.5, MobileNetV3-x0.75, MobileNetV3-x1.0, MobileNetV3-x1.4 \\

\midrule
\multirow{4}{*}{Transformer} & ViT \cite{DBLP:conf/iclr/DosovitskiyB0WZ21} & ViT-B/16, ViT-B/32 \\
                             & DeiT \cite{touvron2020training} & DeiT-Ti, DeiT-S, DeiT-B \\
                             & ViTAE \cite{xu2021vitae} & ViTAE-T, ViTAE-S \\
                             & Swin Transformer \cite{liu2021swin} & Swin-T, Swin-S, Swin-B \\

\midrule
\multirow{1}{*}{MLP} & MLP-Mixer \cite{tolstikhin2021mlp} & Mixer-B/16, Mixer-L/16 \\
\bottomrule
\end{tabular}
}
\end{table*}

\begin{table*}
\parbox{.54\linewidth}{
\centering
\caption{General training techniques.}
\label{tab:weight}
\resizebox{1\linewidth}{!}{
\begin{tabular}{l l l}
\toprule
\textbf{Category} & \textbf{Factors} \\ 
\midrule
Training Supervision & 
Knowledge Distillation~\cite{Hinton2015DistillingTK}, Self-supervised Training~\cite{He_2020_CVPR_ssl} \\
\midrule
Weight Adjustment & 
Weight Averaging \cite{pavel2018ema}, Weight Re-parameterization~\cite{ding2021repvgg} \\
\midrule
Regularization & Label Smoothing~\cite{Hinton2015DistillingTK_lablesmooth}, Dropout~\cite{JMLR:v15:srivastava14a_dropout} \\
\midrule
\multirow{2}{*}{Training Data} &  \multirow{2}{*}{\tabincell{l}{Data Augmentation~\cite{hendrycks2020augmix}\cite{zhang2018mixup}, Large-scale Pre-training~\cite{ILSVRC1521k}, \\Adversarial Training \cite{madry2017towards}}} \\

& \\
\midrule
Optimizer & SGD~\cite{ruder2017overview_sgd}, AdamW~\cite{loshchilov2018decoupled_adamw} \\
\bottomrule
\end{tabular}
}
}
\quad
\parbox{.43\linewidth}{
\centering
\caption{Characteristics of noise types.}
\label{tab:noisetype}
\resizebox{1\linewidth}{!}{
\begin{tabular}{l l l l l l}
\toprule
\multirow{2}{*}{\textbf{Noise Types}} & \multirow{2}{*}{\tabincell{l}{\textbf{Adversarial} \\\textbf{Noise}}} & \multirow{2}{*}{\tabincell{l}{\textbf{Natural} \\\textbf{Noise}}} & \multirow{2}{*}{\tabincell{l}{\textbf{System} \\\textbf{Noise}}} \\
& & & \\
\midrule
Model Dependent & \ding{51} & \ding{55} & \ding{55} \\
\midrule
Human Perceptible & \ding{55} & \ding{51} & \ding{55} \\
\midrule
Occurrence Frequency & Low & High & High \\
\midrule
Image Agnostic & \ding{55} & \ding{51} & \ding{51} \\
\midrule
Computational Cost & High & Middle & Low \\ 
\bottomrule
\end{tabular}
}
}
\end{table*}

\subsubsection{Architecture design}
To conduct a thorough evaluation and explore the robustness trends of different architecture families, our repository tends to cover as many architectures as possible. As for the {\textbf{CNNs}}, we choose the classical large architectures including ResNet-series~(ResNet, ResNeXt, WideResNet) and DenseNet, the lightweight ones including ShuffleNetV2 and MobileNetV2, the reparameterized architecture RepVGG, the NAS models including RegNet, EfficientNet and MobileNetV3, and sub-networks sampled from the BigNAS \cite{yu2020bignas} supernet. As for the recently prevalent {\textbf{Vision Transformers}}, we reproduce ViT, DeiT, ViTAE, and Swin Transformer. Besides, we also include the {\textbf{MLP}} based architecture MLP-Mixer. All the models available are listed in \autoref{tab:architecture}. In total, we collect 49 human-designed off-the-shelf model architectures and sampled 1200 subnet architectures from the supernets. For fair comparisons of robustness, as for all human-designed off-the-shelf models we keep the aligned training technique setting (\eg, the same data augmentation techniques).

\subsubsection{Training techniques}
An increasing number of techniques have been proposed in recent years to train deep learning models with higher accuracy. Some of them have been proved to be influential for model robustness by previous literature~\cite{ali2019label,fu2020label,li2020shape}, \eg, data augmentation. However, with investigations only on limited architectures and noise types, these conclusions might not be able to unveil the intrinsic nature behind. Besides, there are broader training techniques but their relationships with robustness are still ambiguous. To give more exact answers about the influence on robustness, we summarize 10+ mainstream training techniques into five different categories (see \autoref{tab:weight}) and systematically benchmark their effects on robustness. The relevant pre-trained models using different training techniques are open-sourced.


\subsection{Evaluation strategies}

\subsubsection{Dataset}
For all of our experiments, we use the large-scale ImageNet-1K dataset \cite{deng2009imagenet}, which contains 1,000 classes of colored images of size 224 * 224 with 1,281,167 training examples and 50,000 test instances.

\subsubsection{Noise types}
There exist various noise types in the real-world scenarios but most papers only consider a fraction of them (\eg, adversarial noises), which may fail to fully benchmark model robustness. This inspires us to collect diverse noise sources for a more comprehensive robustness profiling. As shown in Table \ref{tab:noisetype}, we categorize the noise format into three types: adversarial noise, natural noise, and system noise. The robustness of a model should take all these noises into consideration.

\textbf{\emph{Adversarial noises.}} This type of noise is designed to attack DNNs. We choose 9 commonly-used adversarial attacks involving while-box attacks and black-box attacks with various $\ell_p$ perturbation types, including: FGSM \cite{goodfellow6572explaining}, PGD-$\ell_{\infty}$ \cite{madry2017towards}, AutoAttack-$\ell_{\infty}$ \cite{croce2020reliable}, MIM-$\ell_{\infty}$ \cite{dong2018boosting}, PGD-$\ell_{2}$ \cite{madry2017towards}, DDN-$\ell_{2}$ \cite{rony2019decoupling}, C\&W-$\ell_{2}$ \cite{carlini2017towards}, PGD-$\ell_{1}$ \cite{madry2017towards} and transfer-based attack \cite{liu2016delving}. Each attack has 3 different perturbation magnitudes (small $\epsilon$, middle $\epsilon$, and large $\epsilon$). Specifically, for $\ell_{1}$ attacks we set large $\epsilon$=1600.0, middle $\epsilon$=400.0 and small $\epsilon$=100.0; for $\ell_{2}$ attacks we set large $\epsilon$=8.0, middle $\epsilon$=2.0 and small $\epsilon$=0.5; for $\ell_{\infty}$ attacks we set large $\epsilon$=8/255, middle $\epsilon$=2/255 and small $\epsilon$=0.5/255. \emph{See Section~\ref{sec:supp_noises-info} and~\ref{sec:supp_noises-setting} in the supplementary materials for more details.}

\textbf{\emph{Natural noises.}} There are many formats of natural noises in the real world, and we choose four typical natural noise datasets (\ie, ImageNet-C, ImageNet-P, ImageNet-A, and ImageNet-O). \emph{See Section~\ref{sec:natural_noise} in the supplementary materials for more details.} 

\textbf{\emph{System noises.}} Besides, there always exists system noises in the training or inference stages for models deployed in the industry. Slight differences in pre-processing operations may cause different model performances. For example, different decoders (\ie, RGB and YUV) or resize modes (\ie, bilinear, nearest, and cubic scheme) would introduce system noises and in turn influence model robustness. Thus, we use the ImageNet-S dataset \cite{wang2021imagenets} to evaluate model robustness for system noises, {which consists of 10 different industrial operations}. \emph{See Section~\ref{sec:supp_imagenet-s} in the supplementary materials for more details.}

\subsubsection{Metrics}
\label{sec:metrics}

\textbf{Adversarial Noise Robustness.}
To evaluate the robustness against specific adversarial attacks, we use Adversarial Robustness (AR), which is based on commonly used Attack Success Rate (ASR) \cite{lin2019nesterov,li2020yet,wu2019skip} and calculated by \emph{1 - ASR} (higher indicates stronger model). To evaluate the adversarial robustness under the union of different attacks, we use the Worst-Case Attack Robustness (WCAR), which indicates the lower bound of adversarial robustness against multiple adversarial attacks (higher indicates stronger model). \emph{The formal definitions are shown in Section~\ref{sec:supp_metric-adv} of the supplementary materials.}

\textbf{Natural Noise Robustness.}
For ImageNet-C, prior works \cite{hendrycks2018benchmarking} usually adopt mean Corruption Error (mCE) to measure model errors, thus we use \emph{1 - mCE} to measure the robustness for ImageNet-C (we call it ImageNet-C Robustness). For ImageNet-P, we follow the mean Flip Probability defined in \cite{hendrycks2018benchmarking} and take the negative of it (Negative mean Flip Probability, NmFP) as our metric. For ImageNet-A, we simply use the classification accuracy of models. For ImageNet-O, we follow \cite{hendrycks2021natural} and use Area Under the Precision-Recall (AUPR). For the above 4 metrics, higher indicates a stronger model. \emph{Formal definitions are shown in Section~\ref{sec:supp_metric-natural} of the supplementary materials.}

\textbf{System Noise Robustness.}
For system noises, we use the accuracy to measure the robustness to different decode and resize modes. To further evaluate the model stability with different decode and resize methods, we use the standard deviation of accuracy across all combinations of these modes and take the negative of it (Negative Standard Deviation, NSD). The higher NSD of a model indicates the better tolerance of facing various combinations of system noises. \emph{See Section~\ref{sec:supp_metric-system} of the supplementary materials for more details.}





\subsection{Framework}

\begin{figure*}[!htb]
	\centering
	\includegraphics[width=1.0\linewidth]{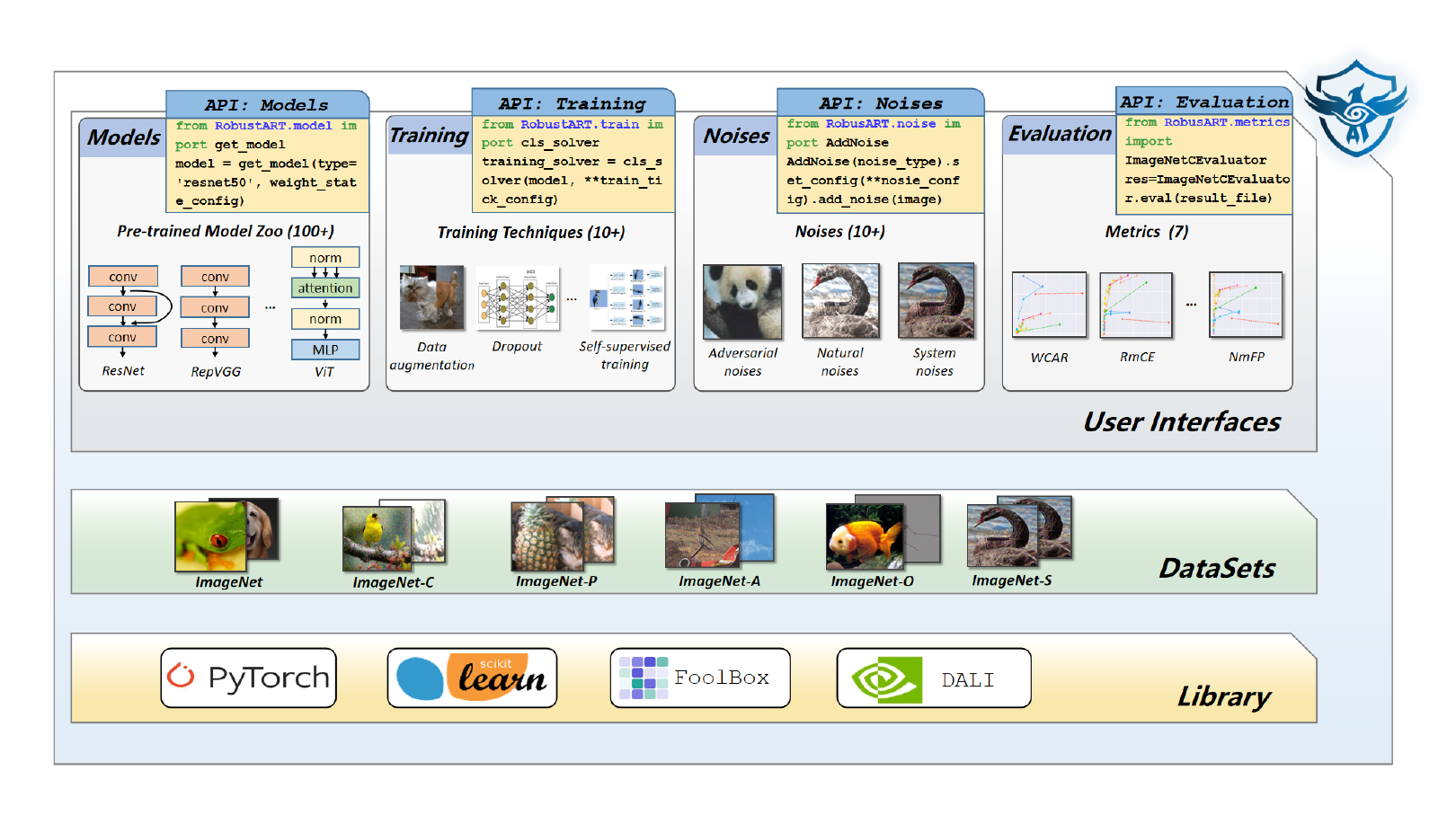}
	\vspace{-0.3in}
	\caption{Our benchmark is built as a modular framework, in which four core modules are provided for users in an easy-to-use way.}
	\label{fig:toolkit}
\end{figure*}

Our benchmark is built as a modular framework, which consists of 4 core modules (Models, Training, Noises, and Evaluation) and provides them in an easy-to-use way (see Figure \ref{fig:toolkit}). All the modules contained are highly extendable for users through API docs on our website. The benchmark is built on Pytorch \cite{paszke2019pytorch} including necessary libraries such as Foolbox \cite{Rauber2017Foolbox} and ART \cite{art2018}.

\textbf{Models.} We have collected 49 human-designed off-the-shelf models and their corresponding checkpoints (100+ pre-trained models in total), including ResNets, ViTs, MLP-Mixers, \etc. Users could flexibly add their model architecture files, load weight configurations, and register their models in the framework to further study the robustness of the customized model architectures.

\textbf{Training.} We provide the implementations and interfaces of all training techniques mentioned in this paper. Users are also able to add customized training techniques to evaluate the robustness. 

\textbf{Noises.} As for adversarial noises we provide 9 adversarial attacks including PGD-$\ell_{\infty}$, AutoAttack-$\ell_{\infty}$, FGSM, PGD-$\ell_{2}$, MIM-$\ell_{\infty}$, PGD-$\ell_{1}$, DDN-$\ell_{2}$, C\&W-$\ell_{2}$ and transfer-based attack; for natural datasets we provide ImageNet-C, ImageNet-P, ImageNet-O, and ImageNet-A; for system noises, we provide ImageNet-S. Users could simply generate corresponding perturbed images through our noise generation APIs. Moreover, users are also recommended to register new noise attack methods.

\textbf{Evaluation.} We provide various evaluation metrics, including WCAR, NmFP, mCE, AUPR, \etc. Users are also recommended to add their own customized metrics through our APIs.

To sum up, based on our open-sourced benchmark, users could (1) conveniently use the source files to reproduce all the proposed results and conduct deeper analyses; (2) add new models, training techniques, noises, and evaluation metrics into the benchmark to conduct additional experiments through our APIs; (3) use our pre-trained checkpoints and research results for other downstream applications or as a baseline for comparisons.

%% file: exp.tex
\section{Experiments and Analyses}
\label{sec:big_experiment}

\subsection{Architecture design towards robustness}
In this section, we first study the influence of architecture design on robustness. In particular, we divide this part into the analysis of human-designed off-the-shelf model architectures, and networks sampled by the neural architecture search.

\subsubsection{Setup}
\label{sec:hancraft_setup}

\qquad\textbf{Human-designed off-the-shelf architectures} To conduct fair and rigorous comparisons among different model architectures, we keep the \emph{\textbf{aligned training techniques}} for each human-designed off-the-shelf architecture. For optimizers, we use SGD \cite{ruder2017overview_sgd} for all model families except for ViTs, DeiTs, ViTAEs, Swin Transformers, and MLP-Mixers; we use AdamW \cite{loshchilov2018decoupled_adamw} for the rest of model families since Transformers and MLP-Mixers are highly sensitive to optimizers (using SGD would cause the failure of training \cite{touvron2020training}). For scheduler, we use cosine scheduler \cite{loshchilov2016sgdr} with maximum training epoch=100 for all models. For data pre-processing, we use standard ImageNet training augmentation, which consists of the random resized crop, random horizontal flip, color jitter, and normalization. \emph{See Section~\ref{sec:supp_handcraft-setting} of the supplementary materials for details.}

\textbf{Architectures sampled from NAS supernets} We choose MobileNetV3, ResNet (basic block architecture), and ResNet (bottleneck block architecture) as the three typical NAS architectures to train the supernets using BigNAS. For optimizer, we use SGD for all supernets; for scheduler, we use the cosine scheduler with maximum training epoch=100; for data pre-processing, we follow the settings for human-designed off-the-shelf architectures and use standard ImageNet training augmentation. For other hyper-parameters, we use label smooth and set batch size=512. \emph{More details are shown in Section~\ref{sec:supp_nas-setting} of the supplementary materials.}

\subsubsection{Human-designed off-the-shelf architectures}
\label{sec:hand-craft_exp}
\begin{figure*}[h]
\includegraphics[width=1.0\linewidth]{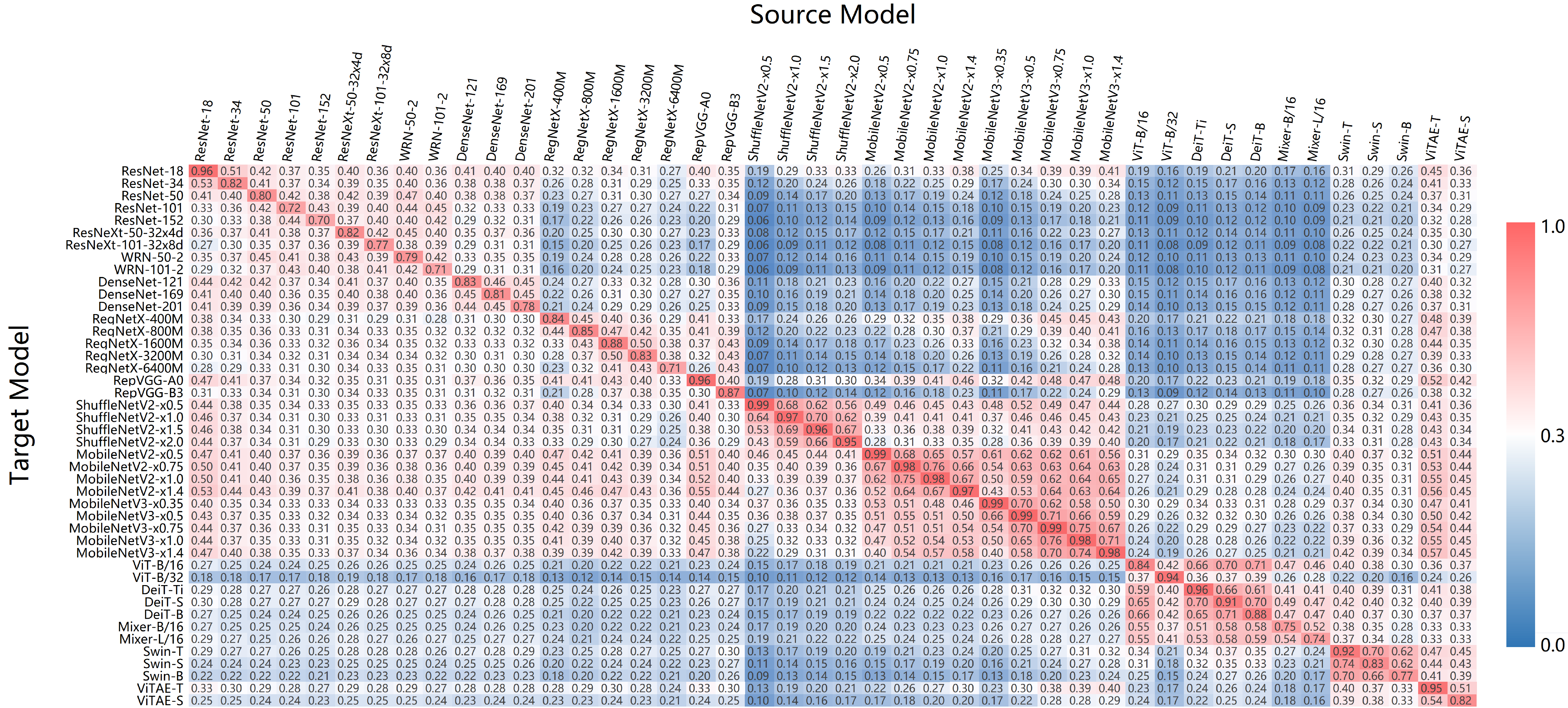}
\caption{Transferability heatmap of human-designed off-the-shelf architectures under FGSM attack, $\epsilon$=8/255. Values mean attack success rates (ASR) from a source model to a target model.}
\label{fig:transfer-rob}
\end{figure*}

To study the effect of human-designed off-the-shelf model architectures on robustness, {we choose 49 models from 15 most commonly used architecture families and keep the \textbf{aligned training settings} for each individual model}. We report the model robustness against different noise types and standard performance (\ie, clean accuracy) \wrt{} model architectures. {We use classical model floating-point operations (FLOPs) and model parameters (Params) to measure model sizes}. \emph{For more results about the robustness and transferability maps with more attack magnitudes, please refer to Figure \ref{fig:supp-handcraft-flops}, \ref{fig:supp-transfer-fgsm2} and \ref{fig:supp-transfer-fgsm05} in the supplementary material.} As shown in Figure \ref{fig:handcraft-rob} and \ref{fig:transfer-rob}, we can draw observations as follows.

\begin{figure*}[h]

\includegraphics[width=1.0\linewidth]{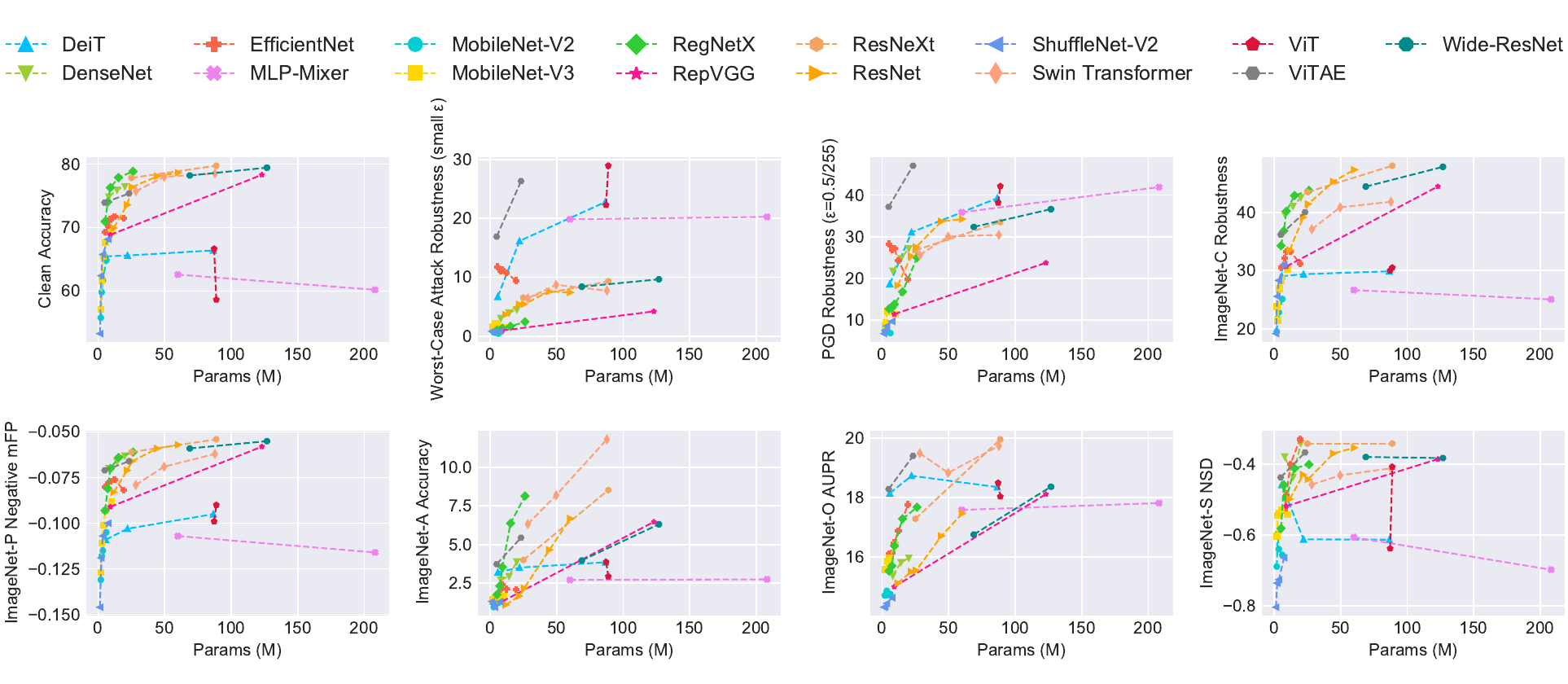}

\caption{Robustness on human-designed off-the-shelf architectures. (\emph{first line}) from left to right: clean accuracy on standard ImageNet, WCAR (small magnitude) under all adversarial attacks, AR under PGD-$\ell_{\infty}$ attack with $\epsilon$=0.5/255, 1-mCE on ImageNet-C; (\emph{second line}) from left to right: NmFP on ImageNet-P, accuracy on ImageNet-A, AUPR on ImageNet-O, and NSD on ImageNet-S. \emph{Results of different FLOPS are in Section~\ref{sec:supp_arch-moreres} of the supplementary materials. Results for different adversarial attacks with different magnitudes are shown in Figure \ref{fig:supp-ddn} to \ref{fig:supp-cw} in the supplementary material.}
}
\label{fig:handcraft-rob}
\end{figure*}

$\blacktriangleright$ \textbf{Model size} ($\textcolor{red}{\uparrow}\textcolor{green}{\downarrow}$\footnote{$\textcolor{red}{\uparrow}$ indicates that this factor has a positive correlation with model robustness, while $\textcolor{green}{\downarrow}$ indicates the opposite.}). For most of the model families (\eg, ResNets, RegNetX), with the increase of the model size (\eg, FLOPs and Params), robustness for adversarial, natural, and system noises improves, and clean accuracy also increases. This indicates the positive influence of model capacity on robustness and task performance within the same model family. However, for some lightweight models such as MobileNetV2, MobileNetV3, and ShuffleNetV2, this conclusion does not entirely hold (\ie, larger models are not necessarily more robust than smaller models). It is even surprising that for EfficientNets, with the increase of the model size, the adversarial robustness decreases. We conjecture it may be due to the increase of input size (for EfficientNets family, the input size monotonically increases from EfficientNet-B0 to B7, \eg, the input size of EfficientNet-B0 is 224, B1 is 240, and B2 is 260). 

$\blacktriangleright$ \textbf{Model architecture} ($\textcolor{red}{\uparrow}\textcolor{green}{\downarrow}$). Based on our large-scale experiments, we can find that the model architecture is a highly significant factor for robustness.

From a coarse-grained perspective, with comparable model sizes and aligned training settings, different model architectures yield significantly different robustness for the same noise. For example, WRN-101-2 and RepVGG-B3 have almost the same Params and clean accuracy performance, whereas their adversarial robustness differs to a large extent (\eg, 9.6 \vs{} 4.1 for worst-case attack robustness, 36.6 \vs{} 23.7 for PGD-$\ell_{\infty}$ robustness with $\epsilon$=0.5/255). 

Specifically, we observe that when compared under the aligned training settings, ViT, ViTAE, and DeiT rank top-3 for worst-case attack robustness; for natural noise robustness, ResNeXt, WideResNet, ResNet rank top-3; for system noises, EfficientNet, DenseNet, ResNeXt rank top-3. Therefore, we could empirically reach the conclusion that \emph{with comparable sizes and aligned training settings, CNNs (e.g., ResNet) perform stronger on natural and system noises, while Transformers and MLP-Mixers are more robust against adversarial noises}. 

Moreover, it is interesting to find that although Swin Transformers are mainly based on the self-attention mechanism like ViTs and DeiTs, their robustness performance is more like CNNs. In other words, in contrast to ViTs, Swin Transformers are comparatively more robust to natural and system robustness, and less robust towards adversarial noises. Some mechanisms in Swin Transformers draw on properties of CNNs, such as window attention (introduce the locality like convolution kernel) and hierarchical architecture, thus we conjecture this might be the reason for the special robustness performance of Swin Transformers. 
We hope all these observations could inspire more in-depth future studies.

$\blacktriangleright$ \textbf{Adversarial transferability}. According to the experimental results, we mainly divide the model architectures into three categories: common CNNs (ResNet, ResNeXt, WideResNet, DenseNet, RegNeXt, and RepVGG), lightweight CNNs (MobileNetV2, MobileNetV3, and ShuffleNetV2), and non-CNNs (ViTs, DeiTs, and MLP-Mixers). In most cases, models in each category are more robust against attacks transferred from other categories while less robust against attacks transferred from themselves. Interestingly, Swin Transformers and ViTAEs show ``moderate'' robustness (\ie, in contrast to other ViTs, Swin Transformers and ViTAEs are more robust against attacks transferred from all other model families, and attacks generated from them also have stronger transferability to all other model families. We speculate that the design of Swin Transformers and ViTAEs has drawn the architecture properties from both CNNs and classical Transformers (\eg, multi-head attention, hierarchical architecture, feature pyramid). 

$\blacktriangleright$ \textbf{Evaluation diversity}. The evaluation diversity plays an important role in evaluating robustness, and different evaluation noises may lead to different robustness evaluation results. For example, different noise types may yield different model robustness (\eg, Swin Transformer > ViT on natural noises, while Swin Transformer < ViT on adversarial noises); for the same noise type, different adversarial attacks may yield different robustness (\eg, ViT > ResNet on AutoAttack, but ViT < ResNet on FGSM); even given the same adversarial attack, different perturbation norms or budgets may still cause different results (\eg, for PGD-$\ell_{\infty}$ attack, DeiT > ResNeXt when $\epsilon$=0.5/255, while DeiT < ResNeXt when $\epsilon$=2.0/255; for EfficientNets under PGD attack, adversarial robustness increases with model size when the perturbation norm is $\ell_{1}$ and $\ell_{2}$, while decreases when the perturbation norm is $\ell_{\infty}$). \emph{See Figure \ref{fig:supp-worstcase} to \ref{fig:supp-cw} in the supplementary materials for more results.}

\subsubsection{Architectures sampled from NAS supernets} 
\begin{figure*}[h]
\includegraphics[width=0.99\linewidth]{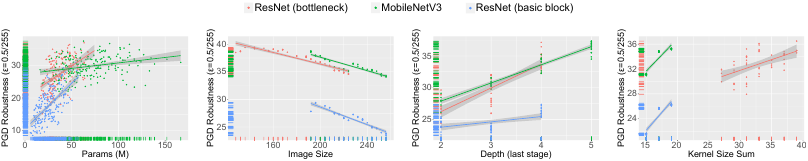}

\caption{Robustness studies on NAS-sampled model architectures using supernets including MobileNetV3 and ResNet families. 
From left to right, we report the effects of model size, input image size, the depth of subnets' last stage, and the sum of convolution kernel sizes on adversarial robustness.
}
\label{fig:nas-rob}
\end{figure*}

We then study the robustness of models sampled from BigNAS \cite{yu2020bignas} supernet. Due to the flexibility of sampled subnets, {we first study the effect of model size on robustness (we totally sampled 600 subnets from 3 supernets) and then dig into the detailed factors that affect robustness in a more fine-grained manner.} Specifically, for each detailed factor (input size, convolution kernel size, model depth, and expand ratio), we fix other model factors and sample 50 subnets from each of the 3 supernets to evaluate robustness (600 subnets in total). As shown in Figure \ref{fig:nas-rob}, we highlight the key findings below. 


$\blacktriangleright$ \textbf{Model size} ($\textcolor{red}{\uparrow}$). Within a specific supernet, increasing model size improves model robustness on adversarial (except for the lightweight supernet MobileNetV3, which is similar to conclusions in the human-designed model study) and natural noises, but there is no apparent effect on system noises. We hypothesize that \emph{for lightweight models, simply increasing their capacities cannot improve robustness.}

$\blacktriangleright$ \textbf{Model depth} ($\textcolor{red}{\uparrow}$). In all supernets, increasing the depth of the last stage (\ie, the number of layers in the subnets' deepest block) could improve the adversarial robustness of the sampled subnets, which shows the deepest stage is of great importance for model robustness in NAS-sampled network. However, for the depth of other stages or the whole network, the relationship is still ambiguous.

$\blacktriangleright$ \textbf{Input size} ($\textcolor{green}{\downarrow}$). In all supernets, when increasing the input size of sampled subnets, their adversarial robustness decreases. This observation is similar to the study on EfficientNets in Sec \ref{sec:hand-craft_exp}, indicating that networks with larger input sizes might be less robust against adversarial attacks. However, the effects of input size on robustness for natural and system noises are ambiguous.

$\blacktriangleright$ \textbf{Convolution kernel size} ($\textcolor{red}{\uparrow}$). For sampled subnets, increasing the total number of convolution kernel sizes in all stages could improve the adversarial robustness of subnets, while the effects on natural and system noises are unclear. For each stages' convolution kernel size, there seems no correlation with model robustness. 

To sum up, we found that the architecture design has a huge impact on robustness (especially the sizes and families). Meanwhile, the noise diversity is highly important to robustness evaluation (slight differences would cause opposite observations), we therefore recommend researchers to use more comprehensive and diverse noises when evaluating model robustness. 


\subsection{Training techniques towards robustness}
\label{sec:weights_exp}

\subsubsection{Setup}
In total, we choose 11 different training techniques, including weight averaging, label smoothing, ImageNet-21K pre-training, Mixup, Augmix, Dropout, weight re-parameterization, knowledge distillation, MOCO v2 self-supervised training, PGD-$\ell_{\infty}$ adversarial training, and AdamW optimizer. In particular, for large-scale pre-training, we adopt ImageNet-21K as the pre-training data; for data augmentation, we select Mixup \cite{zhang2018mixup} and Augmix \cite{hendrycks2020augmix}; for adversarial training, we follow \cite{robustness, xie2019denoising} and adversarially train models with PGD-$\ell_{\infty}$ attacks \cite{madry2017towards}; for self-supervised training, we use MoCo \cite{He_2020_CVPR_ssl}. Due to the huge time consumption of training all models with or without these 11 training techniques on ImageNet, we choose several typical and widely-used architectures for each training technique (\eg, ResNet-50, RegNetX-3200M,  RepVGG-A0, RepVGG-B3, ShuffleNetV2-x2.0, MobileNetV3-x1.4, ViT-B/16, and Mixer-B/16). For each training technique, we either apply or disable this training technique for different model architectures with all other training settings being the same. \emph{We defer more details to Section~\ref{sec:supp_technique-setting} in the supplementary materials.}

\subsubsection{Results and analysis}
\begin{figure*}[h]
\subfigure[adversarial training]{
\includegraphics[width=0.49\linewidth]{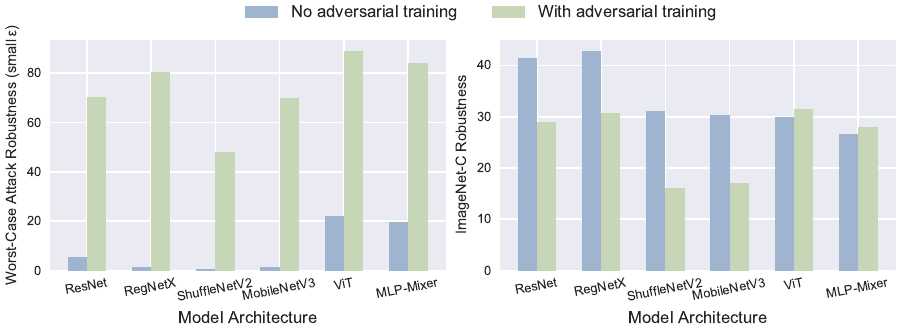}
 \label{fig:weight-0}
}
\subfigure[label smoothing]{
\includegraphics[width=0.49\linewidth]{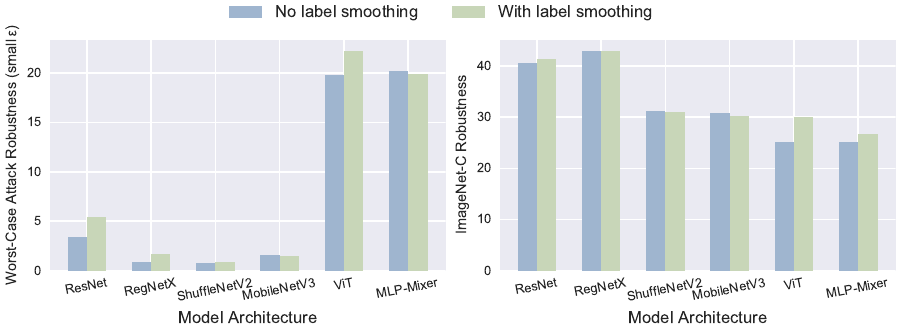}
 \label{fig:weight-1}
}
\subfigure[data augmentation (Mixup)]{
\includegraphics[width=0.49\linewidth]{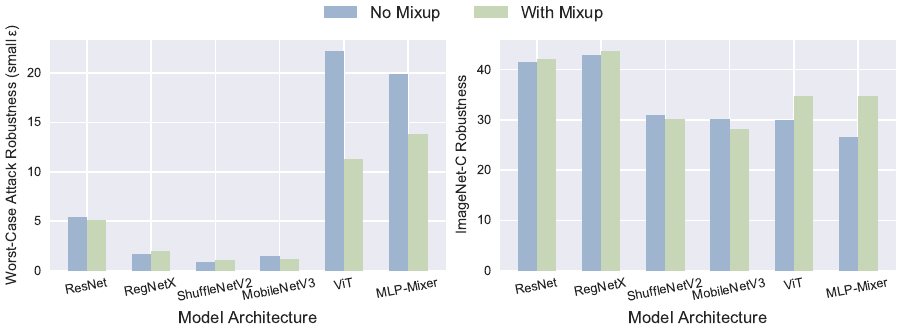}
 \label{fig:weight-2}
}
\subfigure[optimizer]{
\includegraphics[width=0.49\linewidth]{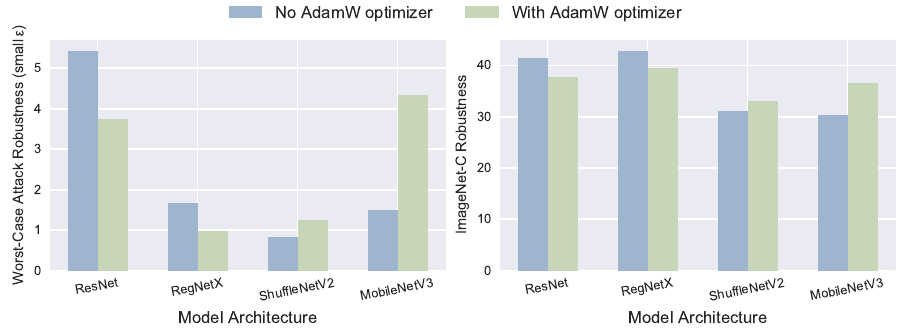}
 \label{fig:weight-3}
}
\subfigure[weight averaging]{
\includegraphics[width=0.49\linewidth]{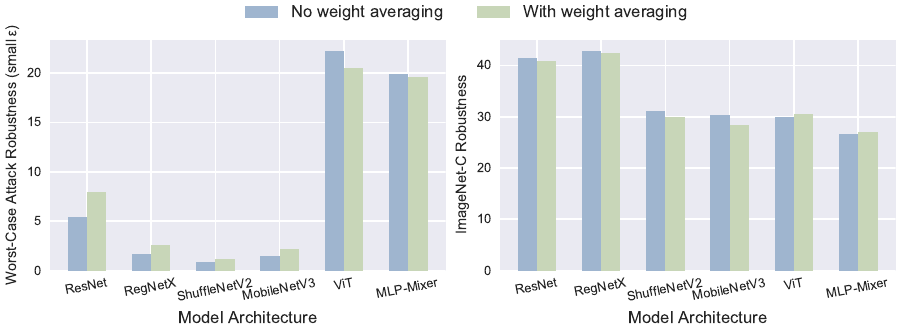}
 \label{fig:weight-4}
}
\subfigure[ImageNet-21K pre-training]{
\includegraphics[width=0.49\linewidth]{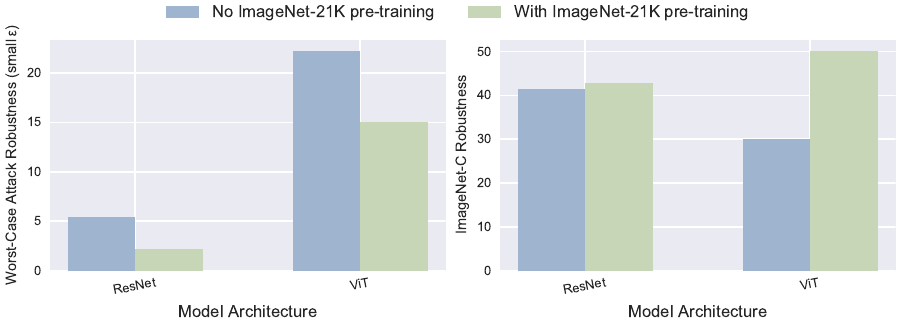}
 \label{fig:weight-5}
 
}

\caption{Robustness of different model architectures with or without the selected training techniques on adversarial and natural noises. \emph{More results about other training techniques are shown in Figure \ref{fig:supp-21kpretrain} to \ref{fig:supp-weightrepara} in Section~\ref{sec:supp_technique-moreres} of the supplementary materials.}} 
\label{fig:weightstate-rob}
\end{figure*}

We list some of the representative results in Figure \ref{fig:weightstate-rob}, and \emph{more results about all training techniques are shown in Figure \ref{fig:supp-21kpretrain} to \ref{fig:supp-weightrepara} in the supplementary materials}. From the experimental results, we draw several observations as follows.

$\blacktriangleright$  \textbf{Training Data}

\quad (1) \textbf{Adversarial training} ($\textcolor{red}{\uparrow}\textcolor{green}{\downarrow}$). For CNNs, adversarial training largely boosts adversarial robustness and robustness against ImageNet-P, but reduces the clean accuracy as well as the robustness against ImageNet-C and system noises. For ViTs and MLP-Mixer, adversarial training also largely improves adversarial robustness and robustness against ImageNet-P, slightly improves the robustness against ImageNet-C and system noises, whereas slightly reduces the clean accuracy (much smaller than the drop on CNNs, \eg, 76.3 $\rightarrow$ 57.8 for ResNet-50, while 66.6 $\rightarrow$ 60.1 for ViT-B/16). 

\quad (2) \textbf{Data augmentation} ($\textcolor{red}{\uparrow}\textcolor{green}{\downarrow}$). For adversarial noises, data augmentation reduces worst-case attack robustness in most cases; for natural noises (ImageNet-C, -P, -A, and -O), data augmentation yields a stronger model in most cases, and the improvement on ViTs and MLP-Mixers is larger than CNNs (\eg, Augmix improves the ImageNet-C robustness of ResNet-50 from 41.4 to 44.3, while improves ImageNet-C robustness of Mixer-B/16 from 26.6 to 40.3); for system noises, Mixup improves robustness while Augmix reduces robustness, showing that different data augmentation techniques might have different impacts on model robustness.

\quad (3) \textbf{ImageNet-21K pre-training} ($\textcolor{red}{\uparrow}\textcolor{green}{\downarrow}$). For both CNNs and ViTs, this technique improves robustness on all of the natural noises and reduces robustness on system noises and adversarial noises (under WCAR metric). And the robustness improvement on ViTs is much bigger than CNNs. (\eg, ImageNet-A robustness of ResNet-50 improves from 2.1 to 8.1, while ViT-B/16 improves from 3.8 to 23.1; ImageNet-C robustness of ResNet-50 improves from 41.4 to 42.7, while ViT-B/16 improves from 29.9 to 50.2)

$\blacktriangleright$  \textbf{Weight Adjustment}

\quad (1) \textbf{Weight averaging} ($\textcolor{red}{\uparrow}\textcolor{green}{\downarrow}$). For CNNs, weight averaging improves the adversarial robustness (especially when perturbation budget is small, \eg, weight averaging improves the robustness of RegNetX-3200M from 19.6 to 24.2 under MIM-$\ell_{\infty}$ attack, $\epsilon$=0.5/255), but slightly reduces the robustness against natural noises. For ViTs and MLP-Mixers, the impact of weight averaging on robustness is comparatively small. 

$\blacktriangleright$  \textbf{Regularization}

\quad (1) \textbf{Label smoothing} ($\textcolor{red}{\uparrow}\textcolor{green}{\downarrow}$). Label smoothing boosts robustness against all adversarial noises in most cases (the improvement is especially large for MLP-Mixer against middle and large perturbation budget, \eg, label smoothing boosts the robustness of Mixer-B/16 from 0.2 to 8.3 under PGD-$\ell_{\infty}$ attack, $\epsilon$=2.0/255), while the effects on natural and system noises are comparatively small.

\quad (2) \textbf{Dropout} ($\textcolor{red}{\uparrow}\textcolor{green}{\downarrow}$). For CNNs, Dropout slightly reduces the adversarial robustness in most cases, and the impact on natural and system noises is ambiguous. For ViTs and MLP-mixers, although Dropout still causes a slight drop in adversarial robustness in most cases, it improves model robustness against all of the natural noises and system noises, and the improvement is large when it comes to ViTs (\eg, Dropout boosts ImageNet-C robustness of ViT-B/16 from 29.9 to 40.5).

$\blacktriangleright$  \textbf{Optimizer}

\quad (1) \textbf{AdamW}($\textcolor{red}{\uparrow}\textcolor{green}{\downarrow}$). 
For common CNNs (ResNet and RegNetX), AdamW reduces robustness against all noises (adversarial, natural, and system) in most cases compared to SGD. However, for lightweight models (MobileNetV3 and ShuffleNetV2), AdamW could improve robustness against all noises as well as clean accuracy. (\eg, For MobileNetV3-x1.4, after using AdamW, clean accuracy improves from 69.2 to 73.6, PGD-$\ell_{\infty}$ robustness improves from 11.4 to 21.6, ImageNet-C robustness improves from 30.2 to 36.5)

For other factors that are not mentioned in the main body, they show no evident effects on robustness. To sum up, there still does not exist an ``once-for-all'' training technique that yields stronger robustness for all architectures and all noises. Some techniques may even pose opposite effects on the same noise type for different architectures (\eg, AdamW improves adversarial robustness of MobileNetV3, while reducing that for ResNet). Thus, for fair comparisons of model robustness, we should align the training techniques.

\subsection{{Discussions}}
 In contrast to existing studies that evaluate adversarial defenses, this paper aims to conduct a comprehensive and fair robustness investigation of different architectures and training techniques. For architectures, we align different model architectures (including CNNs, Transformers, and MLP-Mixers) with the same training settings (without extra training techniques) for a fair comparison, which is often ignored by most existing literature. For training techniques, instead of studying one specific training technique with all its implementation tricks, we summarize and evaluate 10+ training techniques over several architectures to help better understand how training techniques will influence model robustness. Through extensive ablation studies (including eliminating the extra variables with an aligned training setting for comparing different model architectures), we comprehensively demonstrate how different model architectures and training techniques impact the robustness. Based on large-scale experiments and fair comparisons, we have obtained a wide variety of observations. In this section, we offer some of the most important and consistent ones and provide further discussions and analyses on them.

\begin{table*}[t]
\centering
\caption{Robustness studies for adversarial training on ViT-B/16 and Mixer-B/16 models under the aligned training setting. ``Vanilla'' means the model is vanilla trained based on the aligned training setting in \ref{sec:hancraft_setup}. ``+Half AdvTrain'' means adversarial training via Half PGD-$\ell_{\infty}$ Adversarial Training scheme ($\epsilon$=16/255) based on the aligned training setting.}
\label{tab:halfAdvTraining}
\resizebox{1\linewidth}{!}{
\begin{tabular}{llccccccccc}
\toprule
\textbf{Architecture} & \textbf{\begin{tabular}[c]{@{}l@{}}Training\\ Settings\end{tabular}}           & \textbf{\begin{tabular}[c]{@{}c@{}}Clean\\ Accuracy\end{tabular}} & \textbf{\begin{tabular}[c]{@{}c@{}}WCAR\\ (large $\epsilon$)\end{tabular}} & \textbf{\begin{tabular}[c]{@{}c@{}}WCAR\\ (middle $\epsilon$)\end{tabular}} & \textbf{\begin{tabular}[c]{@{}c@{}}WCAR\\ (small $\epsilon$)\end{tabular}} & \textbf{\begin{tabular}[c]{@{}c@{}}ImageNet-A\\ Accuracy\end{tabular}} & \textbf{\begin{tabular}[c]{@{}c@{}}ImageNet-O\\ AUPR\end{tabular}} & \textbf{\begin{tabular}[c]{@{}c@{}}ImageNet-C\\ Robustness\end{tabular}} & \textbf{\begin{tabular}[c]{@{}c@{}}ImageNet-P\\ NmFP\end{tabular}} & \textbf{\begin{tabular}[c]{@{}c@{}}ImageNet-S\\ NSD\end{tabular}} \\ \midrule
ViT-B/16              & \multicolumn{1}{l|}{\begin{tabular}[c]{@{}l@{}}Vanilla\end{tabular}} & 66.662                                                            & 0.000                                                                      & 0.063                                                                       & 22.259                                                                     & 3.866                                                                  & 18.490                                                             & 29.997                                            & -0.099                                                             & -0.652                                                            \\
ViT-B/16              & \multicolumn{1}{l|}{+Half AdvTrain}                                                     & 69.960                                                            & 7.624                                                                      & 22.335                                                                      & 51.703                                                                     & 5.040                                                                  & 20.670                                                             & 42.344                                            & -0.047                                                             & -0.385                                                            \\
Mixer-B/16            & \multicolumn{1}{l|}{\begin{tabular}[c]{@{}l@{}}Vanilla\end{tabular}} & 62.598                                                            & 0.000                                                                      & 0.099                                                                       & 19.857                                                                     & 2.706                                                                  & 17.580                                                             & 26.648                                            & -0.107                                                             & -0.618                                                            \\
Mixer-B/16            & \multicolumn{1}{l|}{+Half AdvTrain}                                                     & 68.296                                                            & 7.151                                                                      & 13.485                                                                      & 32.447                                                                     & 3.853                                                                  & 19.780                                                             & 38.396                                            & -0.065                                                             & -0.407                                                            \\ \bottomrule
\end{tabular}
}
\end{table*}

\begin{table*}[t]
\centering
\caption{Robustness studies for adversarial training on ViT-B/16 under the optimal ViT training setting. ``Vanilla'' means the model is vanilla trained based on the optimal ViT training setting in \cite{touvron2020training}. ``+Half AdvTrain'' means using Half PGD-$\ell_{\infty}$ Adversarial Training scheme ($\epsilon$=0.5/255) based on the optimal ViT training setting.}
\label{tab:optimalHalfAdvTraining}
\resizebox{1\linewidth}{!}{
\begin{tabular}{llccccccccc}
\toprule
\textbf{Architecture} & \textbf{\begin{tabular}[c]{@{}l@{}}Training\\ Settings\end{tabular}} & \textbf{\begin{tabular}[c]{@{}c@{}}Clean\\ Accuracy\end{tabular}} & \textbf{\begin{tabular}[c]{@{}c@{}}WCAR\\ (large $\epsilon$)\end{tabular}} & \textbf{\begin{tabular}[c]{@{}c@{}}WCAR\\ (middle $\epsilon$)\end{tabular}} & \textbf{\begin{tabular}[c]{@{}c@{}}WCAR\\ (small $\epsilon$)\end{tabular}} & \textbf{\begin{tabular}[c]{@{}c@{}}ImageNet-A\\ Accuracy\end{tabular}} & \textbf{\begin{tabular}[c]{@{}c@{}}ImageNet-O\\ AUPR\end{tabular}} & \textbf{\begin{tabular}[c]{@{}c@{}}ImageNet-C\\ Robustness\end{tabular}} & \textbf{\begin{tabular}[c]{@{}c@{}}ImageNet-P\\ NmFP\end{tabular}} & \textbf{\begin{tabular}[c]{@{}c@{}}ImageNet-S\\ NSD\end{tabular}} \\ \midrule
ViT-B/16              & \multicolumn{1}{l|}{Vanilla}                                 & 75.132                                                            & 0.000                                                                          & 0.035                                                                       & 18.693                                                                     & 6.470                                                                  & 20.910                                                             & 41.520                                                                   & -0.066                                                             & -0.409                                                            \\
ViT-B/16              & \multicolumn{1}{l|}{+Half AdvTrain}                                           & 74.684                                                            & 0.002                                                                      & 17.934                                                                      & 75.794                                                                     & 6.573                                                                  & 19.471                                                             & 48.457                                                                   & -0.038                                                             & -0.378                                                            \\ \bottomrule
\end{tabular}
}
\end{table*}

\subsubsection{Adversarial training is universally effective for the robustness of Transformers and MLP-Mixers}
During our analysis for the impact of training techniques on robustness in Sec \ref{sec:weights_exp}, we follow the classical PGD-$\ell_{\infty}$ adversarial training scheme in \cite{robustness, xie2019denoising} and only feed adversarial examples into models in each mini-batch to conduct training. Based on this adversarial training scheme, we find that in contrast to CNNs, adversarially trained ViTs and MLP-Mixers have a smaller clean accuracy drop and larger robustness improvement (even the natural and system robustness are improved), which indicates ViTs and MLP-Mixers perform better with adversarial training than CNNs. 

To further investigate this observation, we follow \cite{zhang2019interpreting} and use another PGD-$\ell_{\infty}$ adversarial training scheme to train models and evaluate their robustness. Specifically, we simultaneously feed clean examples and their corresponding adversarial examples in each mini-batch into models with a ratio of 1:1 (we call this scheme Half Adversarial Training). From the results in Table \ref{tab:halfAdvTraining}, we find that Half Adversarial Training could also largely increase the robustness of Transformers and MLP-Mixers towards all types of noises (\ie, adversarial, natural, and system noises). More importantly, we find that this adversarial training scheme could improve \emph{the accuracy of Transformers and MLP-Mixers on clean examples}. However, Half Adversarial Training also decreases the accuracy of CNNs on clean examples, which indicates that adversarial training can be treated as a special technique for Transformers and MLP-Mixers to universally improve model performance if harnessed properly. \emph{We defer more results about Half Adversarial Training scheme in Figure \ref{fig:supp-halfadvtrain} in the supplementary materials.} 

Note that, for all the above experiments for adversarial training, ViTs are adversarially trained based on our aligned training settings (\ie, without using extra training techniques, \eg, AutoAugment or Dropout). Therefore, we further evaluate the effect of adversarial training for ViTs under the optimal ViT training settings \cite{touvron2020training} (\eg, with Mixup, AutoAugment, \etc.). In particular, we conduct Half Adversarial Training on a ViT-B/16 using the optimal setting and compare its robustness with another vanilla trained ViT-B/16 model under the same optimal setting. From the results in Table \ref{tab:optimalHalfAdvTraining}, we could reach similar observations, namely this adversarial training scheme could largely boost the robustness of ViT on all types of noises. In fact, we also conduct the standard PGD adversarial training (only feed adversarial examples into models in each mini-batch) for ViT-B/16 under the optimal settings, however, we observe that the model cannot converge. We speculate that it is too difficult for the model to fit the augmented adversarial examples since there have been various data augmentations in the optimal ViT training procedure \cite{bai2021transformers}. 

Overall, to the best of our knowledge, we are the first to propose this conclusion through extensive experiments on ImageNet that \emph{adversarial training is universally effective for the robustness of Transformers and MLP-Mixers}. We highly advocate researchers to use adversarial training to robustify their Transformers and MLP-Mixers in practice and further investigate the nature behind.

\subsubsection{Vision transformers are more robust against adversarial noises while less robust against natural noises compared to CNNs under aligned settings}
Some of the recent works \cite{mahmood2021On,naseer2021intriguing,Aldahdooh2021Reveal,shao2021Vit,Paul2021Vision} have already studied the adversarial or natural robustness of vision transformers, and reached various conclusions. For adversarial robustness, \cite{mahmood2021On} shows that individual vision transformers are just as vulnerable as their CNN counterparts to white-box adversaries, while \cite{Aldahdooh2021Reveal} and \cite{shao2021Vit} both demonstrates that ViTs possess better adversarial robustness when compared with CNNs. For robustness against natural noises, \cite{naseer2021intriguing} shows that a ViT with comparable parameters is more robust to image corruptions than the CNN trained with augmentations, and \cite{Paul2021Vision} shows that ViTs are more robust than CNNs under corruptions and OOD distributions.

However, as we can see from the above studies, some of their conclusions and observations are contradictory to others. We attribute this phenomenon to the unfair experimental settings and incomplete noises evaluation. For instance, most of the ViTs used in these works are trained with the standard settings for ViTs which consists of various specially-designed data augmentation techniques (\eg, AutoAugment and RandAugment), while they are not used for CNNs. Studies in \ref{sec:weights_exp} have revealed that data augmentation techniques could reduce adversarial robustness and improve natural robustness in most cases, therefore it will interfere with the correctness of model robustness evaluation when data augmentation techniques are introduced. 

Therefore, in Sec \ref{sec:hand-craft_exp}, we fairly evaluate the robustness of ViTs and other architectures (\eg, CNNs) under the aligned training settings with diverse noises. Through our comprehensive empirical studies, we finally reach the conclusion that under aligned training settings, vision transformers and MLP-Mixers are more robust against adversarial noises while less robust against natural noises compared to CNNs.

\subsubsection{Architecture is highly important for robustness}
Previous works \cite{su2018is, huang2021exploring} have demonstrated that model architecture is a more critical factor to robustness than model size. However, there exist two drawbacks of these works: on the one hand, they do not take the effect of training techniques into account, which may overestimate the robustness of model architectures; on the other hand, the numbers of models and noises used in the evaluation are comparatively small, which make the evaluation incomprehensive. Therefore, this paper keeps the aligned training settings when comparing the robustness of different model architectures, without introducing the influence of training techniques. Through our large-scale studies, we empirically demonstrate the significance of model architecture to robustness, and we hereby provide several interesting phenomena:

(1) Different architectures show different robustness. Generally, CNNs are more robust to natural and system noises while Transformers and MLP-Mixers are more robust to adversarial noises. 

(2) As one of the Transformers architectures, Swin Transformer shows better robustness against natural noises while less robustness against adversarial noises, which is more like CNNs rather than other Transformers. 

(3) Slight modifications in the structure (input size, depth, \etc) of subnets sampled from supernets would affect their adversarial robustness. 

To improve the robustness, besides designing new defense methods, robust architectures could also be taken into consideration. For example, Transformers are more robust for adversarial noises, and CNNs are more robust for natural and system noises. 

\subsubsection{There does not exist a “once-for-all” training technique for improving robustness}
Previous works \cite{kamath2020sgd, fu2020label,hendrycks2020many,hendrycks2020pretrained} have studied the influence of some specific training techniques (\eg, label smoothing) on model robustness. In this work, we summarize and evaluate 10+ training techniques over several architectures to help better understand how training techniques will influence the robustness. We suggest that for fair evaluation, the training settings for different models should be aligned. Although according to our evaluation results, there does not exist a “once-for-all” training technique for improving robustness, we still draw several representative conclusions for researchers to improve their models' robustness which is listed as follows:

(1) When training lightweight networks like ShuffleNetV2, MobileNetV2, and MobileNetV3, we highly recommend researchers use AdamW optimizer instead of SGD. 

(2) When training Transformers or MLP-Mixers, adversarial training is universally effective for model robustness against all types of noises. 

(3) When training MLP-Mixer, label smoothing could largely improve the adversarial robustness.

(4) When training Transformers, we recommend researchers use Dropout, which could largely boost the robustness against natural and system noises. 

\subsubsection{Noise diversity has huge influences on robustness evaluation}
Some studies \cite{carlini2019evaluating} have mentioned the fact that researchers should improve the diversity of attacks when evaluating adversarial defenses. However, there still exists no comprehensive empirical studies to study the effect of noise diversities on robustness evaluation. Through large-scale experiments and analysis, we safely conclude that models demonstrate different robustness facing different noise types (\eg, ViT-B/16 > ResNeXt-101 on AutoAttack, while ViT-B/16 < ResNeXt-101 on ImageNet-C). For the same noise type, different attack methods or norms may also result in different robustness evaluation results (\eg, from EfficientNet-B0 to B4, the robustness against PGD-$\ell_{2}$ increases while the robustness against PGD-$\ell_{\infty}$ decreases). Even for the same attack methods, different perturbation budgets may still influence the robustness evaluation (\eg, as for PGD-$\ell_{\infty}$ attack, DeiTs > ResNeXts when $\epsilon$=0.5/255, while DeiTs < ResNeXts when $\epsilon$=2/255). Thus, we suggest researchers improve noise diversities for a more comprehensive robustness evaluation.


%% file: conclu.tex
\section{Conclusions}

We propose \textbf{RobustART}, the first comprehensive \textbf{Robust}ness benchmark on ImageNet regarding \textbf{AR}chitecture design (49 human-designed off-the-shelf architectures and 1200 networks constructed by the neural architecture search) and \textbf{T}raining techniques (10+ general techniques including training data augmentation) towards diverse noises (adversarial, natural, and system noises). 
In contrast to existing studies that focus on defense methods, RobustART aims to conduct a comprehensive and fair comparison of different architectures and training techniques towards robustness.
Besides discovering several new findings, our benchmark is highlighted for fertilizing the community by providing: (a) an open-source platform for researchers to use and contribute; (b) 100+ pre-trained models publicly available to facilitate robustness evaluation; and (c) new observations to better understand the mechanism towards robust DNN architectures. We welcome community researchers to join us together to continuously contribute to building this ecosystem to better understand deep learning. We will credit contributions from researchers who are not the authors of this paper on the project website.

%% file: supp.tex
\section{Limitations and Broader Impacts}
\label{sec:limit}
RobustART has several limitations, and we list them as follows. (1) Although we have included a large number of architectures and training techniques, due to the rapid emergence of new approaches, there might still exist relevant ones that are not evaluated. (2) We focus on image classification tasks in the first version of our benchmark, and we will continuously develop the benchmark to include more challenging tasks, such as object detection. (3) We presented many intriguing phenomena in our large-scale experiments, but have not dived into some new findings to analyze their causes. We will conduct more thorough studies based on these observations in future work, and we believe that they will also attract broad interests from the community. We will keep the benchmark up-to-date, and hope the community can contribute together to make the ecosystem grows.

Our benchmark will facilitate the studies of deep learning robustness and a better understanding of model vulnerabilities to adversarial examples and other types of noises. We hope that this work will help in building robust model architectures for real-world applications.

\section{Licenses}
\label{sec:licenses}
The code of RobustART is released under Apache License 2.0. Most model architectures are added to the code with the license chosen by the original author. All pre-trained model checkpoints we provided are produced using our RobustART code and are under Apache License 2.0 as well. The ImageNet-1K, ImageNet-21K, ImageNet-A, ImageNet-O, ImageNet-P, ImageNet-S datasets we use are downloaded from the official release. The ImageNet-C datasets we use are generated according to the official code release on GitHub.

\section{Maintenance Plan}
\label{sec:maintenance}
To make our RobustART benchmark energetic and sustainable, we will keep maintaining our benchmark in these following aspects.

\textbf{\emph{Maintain our website and leaderboard.}}
We host our website and leaderboard on \url{http://robust.art}. We will maintain and update our website and leaderboard to make it a sustainable resource center for robustness. At the same time, we will allow other researchers to upload their results on the leaderboard after review.

\textbf{\emph{Maintain our code base.}}
Our code base is an easy-to-use framework that includes the model, training, noises, and evaluation processes. The code of our framework is hosted on GitHub, we will keep maintaining and updating it.

\textbf{\emph{Maintain our pre-trained models.}}
We provide all pre-trained models in our cloud disk, which have taken more than 44GB of disk space. Every pre-trained model is able to download freely. Moreover, we will keep updating pre-trained models with different architectures and training techniques.

What's more, we also plan to expand our benchmark to other tasks (\ie, object detection and semantic segmentation) and other datasets, making our RobustART a robust benchmark across major computer vision tasks and datasets.

\section{Reproducibility and Run Time}
\label{sec:reproducibility}
We provide the code to run this benchmark on GitHub where everyone can download from freely. As for the setup steps and instructions about our code, we provided a detailed document which can be found on our website \url{http://robust.art}. Following the setup page of this document, users can easily install the required run time environment of this codebase.

Besides, there is a 'Get Started' page of this document. Users can experience most of the basic functions including model training, add noise and model evaluation of our code by following the instruction on page.
For pro developers, we also provided a detailed API document. This API document explains the most important Python class and Python method of our code. Other developers can use this document to modify this code for their needs.

Since our benchmark experiments need us to train multiple models and evaluate them on different kinds of datasets, it needs a large amount of GPU resources. The total cost of our GPU resources to build this benchmark is about 20 GPU years. Most of our experiments are run on Nvidia GTX 1080Ti GPU, some experiments such as ImageNet-21K pre-training which needs more computing resources are run on Nvidia Tesla V100 GPU. For one training experiment, we run it on 16 GPUs parallel. For some training experiments such as ViT, we run it on 32 GPUs parallel to avoid out-of-memory problems.

\section{Detailed Information}
\subsection{Datasets}
For all basic experiments we use ImageNet-1K dataset \cite{deng2009imagenet}, which contains 1,000 classes of colored images of size 224 * 224 with 1,281,167 training examples and 50,000 test instances. We believe that different from small data sets such as MNIST and CIFAR-10, ImageNet is more like the application data in the real scene, and thus the robustness evaluation experiments under ImageNet are more valuable and meaningful.

\subsection{Training Techniques}
\textbf{\emph{Knowledge Distillation.}} 
knowledge distillation usually refers to the process of transferring knowledge from a large model (teacher) to a smaller one (student). While large models (such as very deep neural networks or ensembles of many models) have higher knowledge capacity than small models. In our experiments, we train a ResNet-18 model with a pre-trained ResNet-50 as a teacher for knowledge distillation.

\textbf{\emph{Self-Supervised Training.}}
Self-Supervised Training is proposed for utilizing unlabeled data with the success of supervised learning. In our experiments, we use MoCo v2 which is a classic momentum contrast self-supervised learning algorithm.

\textbf{\emph{Weight Averaging.}} 
Stochastic Weight Averaging is an optimization procedure that averages multiple points along the trajectory of SGD, with a cyclical or constant learning rate.
Weight averaging will lead to better results than standard training. It was proved that weight averaging notably improves training of many state-of-the-art deep neural networks over a range of consequential benchmarks, with essentially no overhead.

\textbf{\emph{Weight Re-parameterization.}} 
RepVGG realizes a decoupling of the training-time and inference-time architecture by structural re-parameterization technique so that the inference can be faster than training~\cite{ding2021repvgg}. We set experiment of RepVGG with and without using re-parameterization to find out its influence on robustness.

\textbf{\emph{Label smoothing.}} 
Label Smoothing is a regularization technique that introduces noise for the labels. This accounts for the fact that datasets may have mistakes in them. Assume for a small constant $\epsilon$ the training set label $y$ is correct with probability $1-\epsilon$ and incorrect otherwise. Label Smoothing regularizes a model based on a softmax with $k$ output values by replacing the hard 0 and 1 classification targets with targets of $\frac{\epsilon}{k-1}$ and $1-\epsilon$ respectively.

\textbf{\emph{Dropout.}} 
Dropout is a regularization technique for neural networks that drops a unit (along with connections) at training time with a specified probability. To find out the influence of dropout on robustness, we trained and tested the models with and without dropout.

\textbf{\emph{Data Augmentation.}}
The performance of neural networks often improves with the amount of training data. Data augmentation is a technique to artificially create new training data from existing training data. We use Mixup and Augmix as two kinds of data augmentation methods to find out their influences.

\textbf{\emph{Large-Scale Pre-training.}}
It has been proved that pre-training using a large-scale dataset can improve the performance of neural networks. To find out the influence of large-scale dataset, we set up experiments using pre-training with ImageNet-21K dataset, which consists of 14,197,122 images (about 14 times larger than ImageNet-1K), each tagged in a single-label fashion by one of 21,841 possible classes. 

\textbf{\emph{Adversarial Training.}}
Adversarial training is one of the most effective defense methods towards adversarial noises. During training, it generates adversarial examples using some attacks and then feed these adversarial examples into the model to calculate the loss function and update model weights. The most commonly used adversarial training method is PGD adversarial training, which uses PGD attack to generate adversarial examples during training procedure. 

\textbf{\emph{SGD and AdamW.}}
Optimizers are algorithms or methods used to minimize the loss function or to maximize the efficiency of production. To Find out the influence of different optimizers, we use two kinds of commonly used optimizers in our experiments. Stochastic Gradient Descent (SGD) is the most basic form of gradient descent. SGD subtracts the gradient multiplied by the learning rate from the weights. Despite its simplicity, SGD has strong theoretical foundations. AdamW is a stochastic optimization method that modifies the typical implementation of weight decay in Adam, by decoupling weight decay from the gradient update.

\subsection{Noises}
\label{sec:supp_noises-info}
\subsubsection{Adversarial noises}
\textbf{\emph{Fast Gradient Sign Method (FGSM).}} The FGSM \cite{goodfellow6572explaining} attack is an early-proposed one-step adversarial attack with $\ell_{\infty}$ norm. It simply substitutes the class variables with the target type of smallest recognition probability in the adversarial perturbation, and then subtracts the perturbation from the original images. Although it yields comparably weak attack ability, the main characteristic of FGSM is its low computational cost. 

\textbf{\emph{Projected Gradient Descent (PGD).}} The PGD \cite{madry2017towards} attack is one of the most powerful first-order attacks in the adversarial literature. Instead of a single step of calculating gradients and doing the gradient descent, it iterates through multiple steps with a gradient projection after each step. With different perturbation norm types, there are mainly PGD-$\ell_{1}$, PGD-$\ell_{2}$ and PGD-$\ell_{\infty}$ attack. We use all of them to conduct a comprehensive robustness evaluation. 

\textbf{\emph{Momentum Iterative Method (MIM).}} The MIM \cite{dong2018boosting} is an iterative adversarial attack based on PGD. It integrates the momentum term into the iterative process for attacks, which stabilizes update directions and can escape from poor local maxima during the iterations, resulting in more powerful attack ability and transferability. We use the widely considered MIM-$\ell_{\infty}$ in our evaluations. 

\textbf{\emph{Decoupled Direction and Norm attack (DDN).}} DDN \cite{rony2019decoupling} is an efficient $\ell_{2}$ norm adversarial attack. It optimizes cross-entropy loss and then changes the norm based on whether the sample is adversarial or not. Using this approach to decouple the direction and norm of the adversarial noise leads to fewer attack iterations, thus speed up the generation of adversarial examples. 

\textbf{\emph{AutoAttack.}} The AutoAttack \cite{croce2020reliable} proposes two extensions of the PGD-attack: APGD-CE and APGD-DLR, which overcome failures due to suboptimal step size and problems of the objective function. It then combines the two novel attacks with two complementary existing attacks (white-box FAB attack \cite{croce2020fab} and black-box square attack \cite{ACFH2020square}) to form a parameter-free and computationally affordable attack. AutoAttack is powerful on many classifiers with various defense methods, thus becoming a good choice for evaluating model robustness in adversarial literature. 

\textbf{\emph{Carlini\&Wagner attack (C\&W).}} The C\&W \cite{carlini2017towards} attack is a family of optimization-based attacks for finding adversarial perturbations that minimize the given loss function. It proposes to transform a general constrained optimization problem into an unconstrained optimization formulation using an empirically chosen loss function. The C\&W attack enables strong attack ability, while the whole optimization procedure is time-consuming.  

\subsubsection{Natural noises}
\label{sec:natural_noise}
\textbf{\emph{ImageNet-A.}}
ImageNet-A is a dataset of real-world adversarially filtered images that fool current ImageNet classifiers. To build this dataset, they first download numerous images related to an ImageNet class. Thereafter they delete the images that fixed ResNet-50 classifiers correctly predict. With the remaining incorrectly classified images, they manually select visually clear images and make this dataset~\cite{hendrycks2021natural}.

\textbf{\emph{ImageNet-O.}} 
ImageNet-O is a dataset of adversarially filtered examples for ImageNet out-of-distribution detectors. To create this dataset, they download ImageNet-22K and delete examples from ImageNet-1K. With the remaining ImageNet-22K examples that do not belong to ImageNet-1K classes, they keep examples that are classified by a ResNet-50 as an ImageNet-1K class with high confidence~\cite{hendrycks2021natural}.

\textbf{\emph{ImageNet-C.}} 
The ImageNet-C dataset consists of 15 diverse corruption types applied to validation images of ImageNet. The corruptions are drawn from four main categories: noise, blur, weather, and digital. Each corruption type has five levels of severity since corruptions can manifest themselves at varying intensities. We use all four categories and five levels of severity of this dataset in our benchmark~\cite{hendrycks2018benchmarking}.

\textbf{\emph{ImageNet-P.}} 
Like ImageNet-C, ImageNet-P consists of noise, blur, weather, and digital distortions. The dataset also has validation perturbations and difficulty levels. ImageNet-P departs from ImageNet-C by having perturbation sequences generated from each ImageNet validation image. Each sequence contains more than 30 frames~\cite{hendrycks2018benchmarking}.

\subsubsection{System noises}
\label{sec:supp_imagenet-s}
The ImageNet-S dataset~\cite{wang2021imagenets} consists of 3 commonly used decoder types and 7 commonly used resize types. For the decoder, it includes the implementation from Pillow, OpenCV, and FFmpeg. For resize operation, it includes nearest, cubic, hamming, lanczos, area, box, and bilinear interpolation modes from OpenCV and Pillow tools. 
 
During the process of evaluation, Pillow bilinear mode is set as the default resize method when testing the decoding robustness. This setting is also the default training setting in PyTorch~\cite{paszke2017automatic} official code example. 
Similarly, Pillow is set as the default image decoding tool (same as PyTorch). 

This dataset provides a validation set of ImageNet with different decoding and resize methods, and saves each image file after decoding and resizing it as a $3 \times width \times height$ matrix in a .npy file instead of JPEG. According to the commonly used transform on ImageNet~\cite{deng2009imagenet} test set, it implements pre-processing for images. This dataset provides a matrix of an image after the process of resizing to $3 \times 256 \times 256$ then applied a center crop to $3 \times 224 \times 224$.

\subsection{Metrics}
\subsubsection{Metrics for adversarial noises}
\label{sec:supp_metric-adv}
\textbf{\emph{Adversarial Robustness (AR).}} As widely used in the literature \cite{lin2019nesterov,li2020yet,wu2019skip}, we choose Attack Success Rate~(ASR) as the base evaluation metric. Considering that ASR mainly measures the attack effectiveness while we are about to measure model robustness, we simply modify it to get our adversarial robustness metric: 

\begin{small}
\begin{equation}
\begin{split}
AR(f, &a, \mathbb{D})= \\
&\frac{\sum\limits_{(\mathbf{x}_{i}, \mathbf{y}_{i} \sim \mathbb{D})}^{} \mathbf{1}\left(f\left(\mathbf{x}_{i}\right)=\mathbf{y}_{i}\right) \cdot \mathbf{1}\left(f\left(a\left(\mathbf{x}_{i}, \mathbf{y}_{i}, f\right)\right)=\mathbf{y}_{i}\right)}{\sum\limits_{(\mathbf{x}_{i}, \mathbf{y}_{i} \sim \mathbb{D})}^{} \mathbf{1}\left(f\left(\mathbf{x}_{i}\right)=\mathbf{y}_{i}\right)}
\end{split}
\end{equation}
\end{small}

where \emph{f} is the threat model, \emph{a} is a certain adversarial attack like PGD-$\ell_{\infty}$ that receives input images, label and threat model as input and returns the perturbed images, and $\mathbb{D}$ is the clean dataset. $\mathbf{1}(\cdot)$ is the indicator function. This metric can effectively reveal the relevant accuracy drop of a specific model when facing noises with considering the original clean accuracy of the model. However, it is not enough to help us figure out the deeper conclusion of the model inherent robustness. Thus we design some new metrics.

\textbf{\emph{Worst-Case Attack Robustness (WCAR).}} To aggregate the model adversarial robustness results under different attacks, we follow instructions in \cite{carlini2019evaluating} and choose to use the per-example worst-case attack robustness: 

\begin{small}
\begin{equation}
\begin{split}
WC&AR(f, \mathbb{A}, \mathbb{D})= \\
&\frac{\sum\limits_{(\mathbf{x}_{i}, \mathbf{y}_{i} \sim \mathbb{D})}^{} \left\{\mathbf{1}\left(f\left(\mathbf{x}_{i}\right)=\mathbf{y}_{i}\right) \cdot \prod\limits_{a \in \mathbb{A}}\mathbf{1}\left(f\left(a\left(\mathbf{x}_{i}, \mathbf{y}_{i}, f\right)\right)=\mathbf{y}_{i}\right)\right\}}{\sum\limits_{(\mathbf{x}_{i}, \mathbf{y}_{i} \sim \mathbb{D})}^{} \mathbf{1}\left(f\left(\mathbf{x}_{i}\right)=\mathbf{y}_{i}\right)}
\end{split}
\end{equation}
\end{small}

where \emph{f} is the threat model, $\mathbb{A}$ is the attack set consisting of many different adversarial attacks, and $\mathbb{D}$ is the clean dataset. $\mathbf{1}(\cdot)$ is the indicator function. WCAR means the lower bound adversarial robustness of a model under multiple different adversarial attacks, which is a better metric for aggregating the results of different adversarial attacks appropriately compared to some other methods like simple averaging. Further, we define three aggregating magnitudes of WCAR: WCAR (small $\epsilon$) means the WCAR under all adversarial attacks with small $\epsilon$, WCAR (middle $\epsilon$) means the WCAR under all adversarial attacks with middle $\epsilon$, and WCAR (large $\epsilon$) means the WCAR under all adversarial attacks with large $\epsilon$.

\subsubsection{Metrics for natural noises}
\label{sec:supp_metric-natural}
\textbf{\emph{Top-1 Accuracy.}}
We use top-1 accuracy as the metric for ImageNet-A dataset. It can reflect how good performance of this model when facing the images which fool commonly used ResNet-50.

\textbf{\emph{1-mCE.}}
We use 1-mCE as the metrics for ImageNet-C dataset. Firstly, we compute the top-1 accuracy of on each corruption type $c$ at each level of severity $s$. And then compute top-1 error rate denote by $E_{s,c}^f$. After that we can compute the Corruption Error, with the formula $\mathrm{CE}_{c}^{f}=\left(\sum_{s=1}^{5} E_{s, c}^{f}\right) /\left(\sum_{s=1}^{5} E_{s, c}^{\mathrm{Alex} \mathrm{Net}}\right)$~\cite{hendrycks2018benchmarking}. Now we can average the 15 Corruption Error values and get $mean CE$ or $mCE$ for shot. Finally, we can compute $1 - mCE$ to evlauate ImageNet-C robustness.

\textbf{\emph{AUPR.}}
We use the area under the precision-recall curve (AUPR) as the metrics for ImageNet-O dataset~\cite{hendrycks2021natural}. This metric requires anomaly scores. Our anomaly score is the negative of the maximum softmax probabilities from a model that can classify the 200 ImageNet-O classes.

\textbf{\emph{NmFP.}}
Denote $m$ perturbation sequences with $\mathcal{S}=\left\{\left(x_{1}^{(i)}, x_{2}^{(i)}, \ldots, x_{n}^{(i)}\right)\right\}_{i=1}^{m}$ where each sequence is made with perturbation $p$ The “Flip Probability” of network $f: \mathcal{X} \rightarrow\{1,2, \ldots, 1000\}$ on perturbation sequences $\mathcal{S}$ is

\begin{equation}
\begin{split}
   \mathrm{FP}_{p}^{f}&=\frac{1}{m(n-1)} \sum_{i=1}^{m} \sum_{j=2}^{n} \mathbb{1}\left(f\left(x_{j}^{(i)}\right) \neq f\left(x_{j-1}^{(i)}\right)\right) \\ 
   &=\mathbb{P}_{x \sim \mathcal{S}}\left(f\left(x_{j}\right) \neq f\left(x_{j-1}\right)\right) 
\end{split}
\end{equation}

For noise perturbation sequences, which are not temporally related, $x_1^{(i)}$ is clean and $x_j^{(i)}$ ($j>1$) are perturbed images of $x_1^{(i)}$ Then the $FP$ formula can be recast as $\mathrm{FP}_{p}^{f}=\frac{1}{m(n-1)} \sum_{i=1}^{m} \sum_{j=2}^{n} \mathbb{1}\left(f\left(x_{j}^{(i)}\right) \neq f\left(x_{1}^{(i)}\right)\right)=\mathbb{P}_{x \sim \mathcal{S}}\left(f\left(x_{j}\right) \neq f\left(x_{1}\right) \mid j>1\right)$. Averaging the Flip Probability across all perturbations and get its negative number yields the Negative mean Flip Probability or NmFP~\cite{hendrycks2018benchmarking}.

\subsubsection{Metrics for system noises}
\label{sec:supp_metric-system}
\textbf{\emph{NSD.}}
We use Negative Standard Deviation (NSD) as the metrics for ImageNet-S system noise, which is the negative value of standard deviation across all accuracy on different decoders and resize methods. The formula of it can be written as $NSD = -\sigma(A)$, where $A = \{a_{decoder}^{resize~method}\}$. We use standard deviation because we want to know the stability of a model facing different decoders and resize methods, and we take the negative value of it since we want this value to increase with this model's performance just like other metrics of this benchmark. 

\section{Experimental Setup}
\label{sec:exp_setup}
Here, we provide the details of the experimental settings of our robustness evaluation benchmark.

\subsection{Settings for Architecture Design}
\subsubsection{Human-designed off-the-shelf architectures}
\label{sec:supp_handcraft-setting}
To conduct fair and rigorous comparisons among different models, we keep the aligned training techniques as much as possible for each human-designed off-the-shelf architecture. For optimizer, we use SGD \cite{ruder2017overview_sgd} optimizer with nesterov momentum=0.9 and weight decay=0.0001 for all model families except for ViTs, DeiTs, Swin Transformers, ViTAEs and MLP-Mixers; instead for these five model families, we use AdamW \cite{loshchilov2018decoupled_adamw} with weight decay=0.05 since Transformers and MLP-Mixers are highly sensitive to optimizers, using SGD would cause the failure of training \cite{touvron2020training}. For scheduler, we use cosine scheduler \cite{loshchilov2016sgdr} with maximum training epoch=100 for all models, for models except for ViTs, DeiTs, Swin Transformers, ViTAEs and MLP-Mixers we use base learning rate (lr)=0.1, warm up lr=0.4, and minimal lr=0.0, and for these five model families we use a much smaller learning rate. For data pre-processing, we use standard ImageNet training augmentation, which consists of random resized crop, random horizontal flip, color jitter, and normalization, for all models including Transformers and CNNs. For all models, we also follow the common settings in network training \cite{szegedy2017inception,wolf2020transformers,chen2020robust,gupta2019stochastic} and use label smooth with $\epsilon$=0.1. For other settings, we set batch size=512, the number of loading workers=4, and enable the pin memory. 

\subsubsection{Architectures sampled from NAS supernets}
\label{sec:supp_nas-setting}
As for architectures sampled from NAS supernets, we choose MobileNetV3, ResNet (basic block architecture), ResNet (bottleneck block architecture) as three typical NAS architectures to train supernets using BigNAS. During supernet training, each batch we sample 4 subnets from supernet to calculate loss and update network weight. For optimizer, we use SGD for all supernets with nesterov momentum=0.9 and weight decay. For scheduler, we use cosine scheduler with maximum training epoch=100. For data pre-processing, we follow the settings for human-designed off-the-shelf architectures and use standard ImageNet training augmentation consisting of random resized crop, random horizontal flip, color jitter, and normalization. For other hyper-parameters, we use label smooth with $\epsilon$=0.1 and batch size=512.

When we study the model size towards robustness, we randomly sample 200 subnets without any factor fixing from each supernet and evaluate subnets' robustness towards adversarial, natural, and system noises. When we study the influence of a certain factor (\ie, input size, convolution kernel size, model depth, or expand ratio), we first fix all other factors, ensuring they are the same in all sampled subnets, then we randomly sample 50 subnets and evaluate their robustness. 

\subsection{Settings for Training Techniques}
\label{sec:supp_technique-setting}
In total, we choose 11 different training techniques, including weight averaging, label smoothing, ImageNet-21K pre-training, Mixup, Augmix, Dropout, weight re-parameterization, knowledge distillation, MOCO v2 self-supervised training, PGD-$\ell_{\infty}$ adversarial training, and AdamW optimizer. Due to the huge time consumption of training all 49 human-designed off-the-shelf architectures with or without these 11 training techniques and evaluating their robustness under various noises on ImageNet, we only choose several architectures for each training technique. For Mixup, Augmix, weight averaging, label smoothing, and PGD-$\ell_{\infty}$ adversarial training, we choose ResNet-50, RegNetX-3200M, ShuffleNetV2-x2.0, MobileNetV3-x1.4, ViT-B/16 and Mixer-B/16, covering large CNNs, light-weight CNNs, and ViTs. For knowledge distillation, we only choose ResNet-18 due to time consumption. For MOCO v2 self-supervised training, we choose ResNet-50 since it is the most commonly used architecture in self-supervised training. For ImageNet-21K pre-training we only choose ResNet-50 and ViT-B/16 due to time consumption. For AdamW optimizer, we choose ResNet-50, RegNetX-3200M, ShuffleNetV2-x2.0 and MobileNetV3-x1.4. For weight re-parameterization, we choose RepVGG-A0 and RepVGG-B3 since this technique is well supported by this model architecture. For Dropout, we choose EfficientNet-B0, MobileNetV3-x1.4, ViT-B/16, and Mixer-B/16. 

For the chosen networks of each training technique, We select to apply or disable this training technique during training with all other training settings being the same. For weight averaging, we set decay=0.9999. For label smoothing, we set $\epsilon$=0.1. For ImageNet-21K pre-training, we train the model on ImageNet-21K dataset for 100 epochs, then finetune it on ImageNet-1K for 30 epochs with a small learning rate (0.01 base learning rate and 0.04 warm up learning rate). For Mixup, we set $\alpha$=0.2. For Augmix, we use the same default config as the original paper. For Dropout, we train one kind of model with the default dropout rate while still train the other kind of model without dropout (i.e. set the dropout rate to 0). For weight re-parameterization, we use the model of RepVGG-A0 and RepVGG-B3 with and without weight re-parameterization respectively. The process of weight re-parameterization is done following by the guide of paper of RepVGG~\cite{ding2021repvgg}.  For knowledge distillation, we use the baseline ResNet-50 model as the teacher model, and the ResNet-18 model as the student model. We train it on ImageNet-1K for 100 epochs. For MOCO v2 self-supervised training, we first train it using the unsupervised method for 200 epochs, then finetune it for 100 epochs. For PGD-$\ell_{\infty}$ adversarial training, we set attack iteration=20, step size=0.002 and $\epsilon$=8/255. For AdamW optimizer, we use cosine decay learning rate schedule setting, base learning rate as $1e-5$ and warm up learning rate as $5e-4$. 

\subsection{Settings for Noises}
\label{sec:supp_noises-setting}
\textbf{\emph{Adversarial noises.}}
For all adversarial attacks, we use three different perturbation magnitudes (small, middle and large). Specifically, for $\ell_{1}$ attacks we set large $\epsilon$=1600.0, middle $\epsilon$=400.0 and small $\epsilon$=100.0; for $\ell_{2}$ attacks we set large $\epsilon$=8.0, middle $\epsilon$=2.0 and small $\epsilon$=0.5; for $\ell_{\infty}$ attacks we set large $\epsilon$=8/255, middle $\epsilon$=2/255 and small $\epsilon$=0.5/255. As for other hyper-parameters, for PGD-$\ell_{1}$, PGD-$\ell_{2}$ and PGD-$\ell_{\infty}$ attack we set attack iteration=20, step size=$\frac{3\epsilon}{40}$ and enable random restart; for MIM-$\ell_{\infty}$ we set attack iteration=20, step size=$\frac{3\epsilon}{40}$, and decay factor=1.0; for DDN-$\ell_{2}$ we set optimization steps=10 and factor $\gamma$=0.05; for AutoAttack-$\ell_{\infty}$ we simply follow its original hyper-parameters since it is a parameter-free attack; for C\&W attack we set binary search steps=5, optimization steps=100 and optimization step size=0.001.

\textbf{\emph{Natural noises.}}
For natural noises, we simply follow the standard settings in ImageNet-C, ImageNet-P, ImageNet-A, and ImageNet-O datasets.

\textbf{\emph{System noises.}}
For system noises, we use the standard setting of ImageNet-S dataset, which can be found in Section~\ref{sec:supp_imagenet-s}.

\section{Additional Results}
\subsection{Architecture Design Towards Robustness}
\label{sec:supp_arch-moreres}
\begin{figure*}[h]
\vspace{-0.05in}
\includegraphics[width=1.0\linewidth]{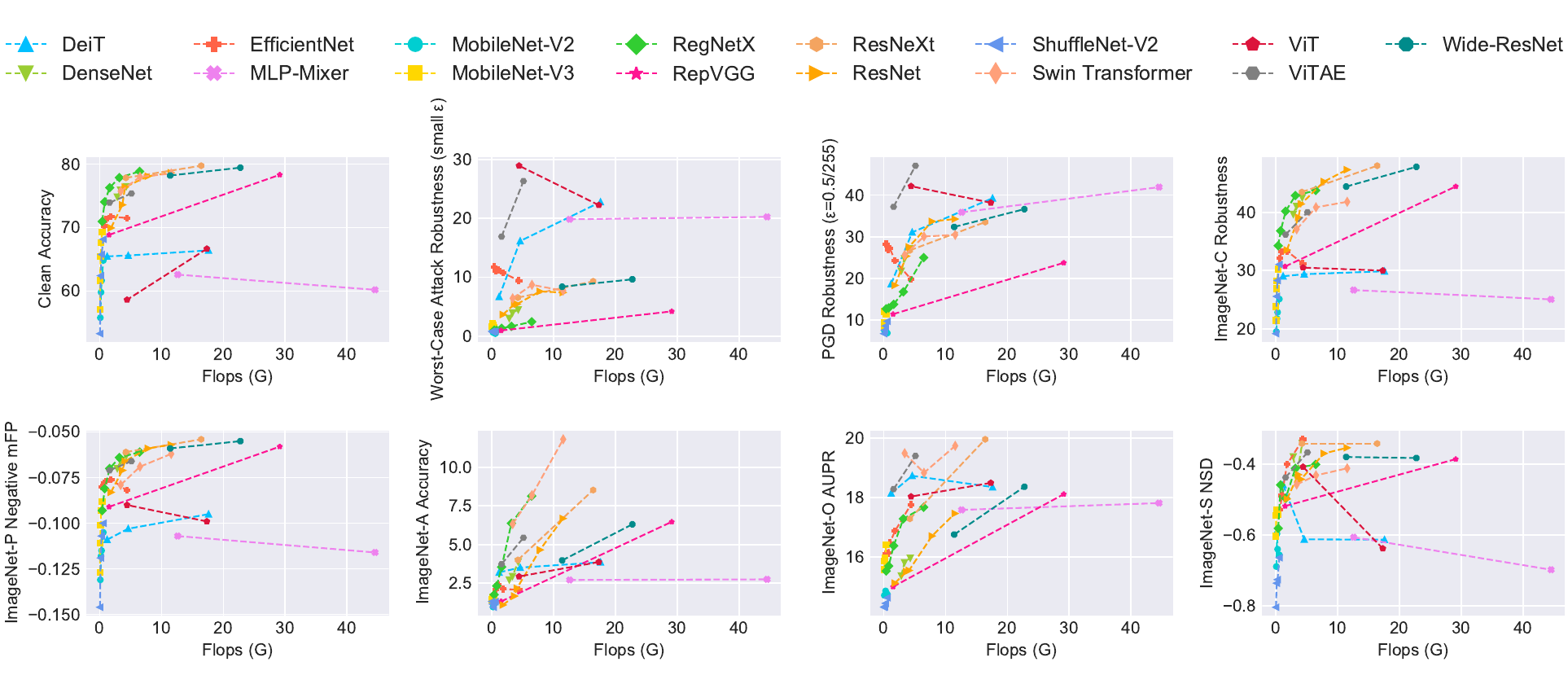}
\caption{Robustness on human-designed off-the-shelf architectures under the view of FLOPs. 
}
\label{fig:supp-handcraft-flops}
\end{figure*}

\begin{figure*}[h]
\vspace{-0.2in}
\includegraphics[width=1.0\linewidth]{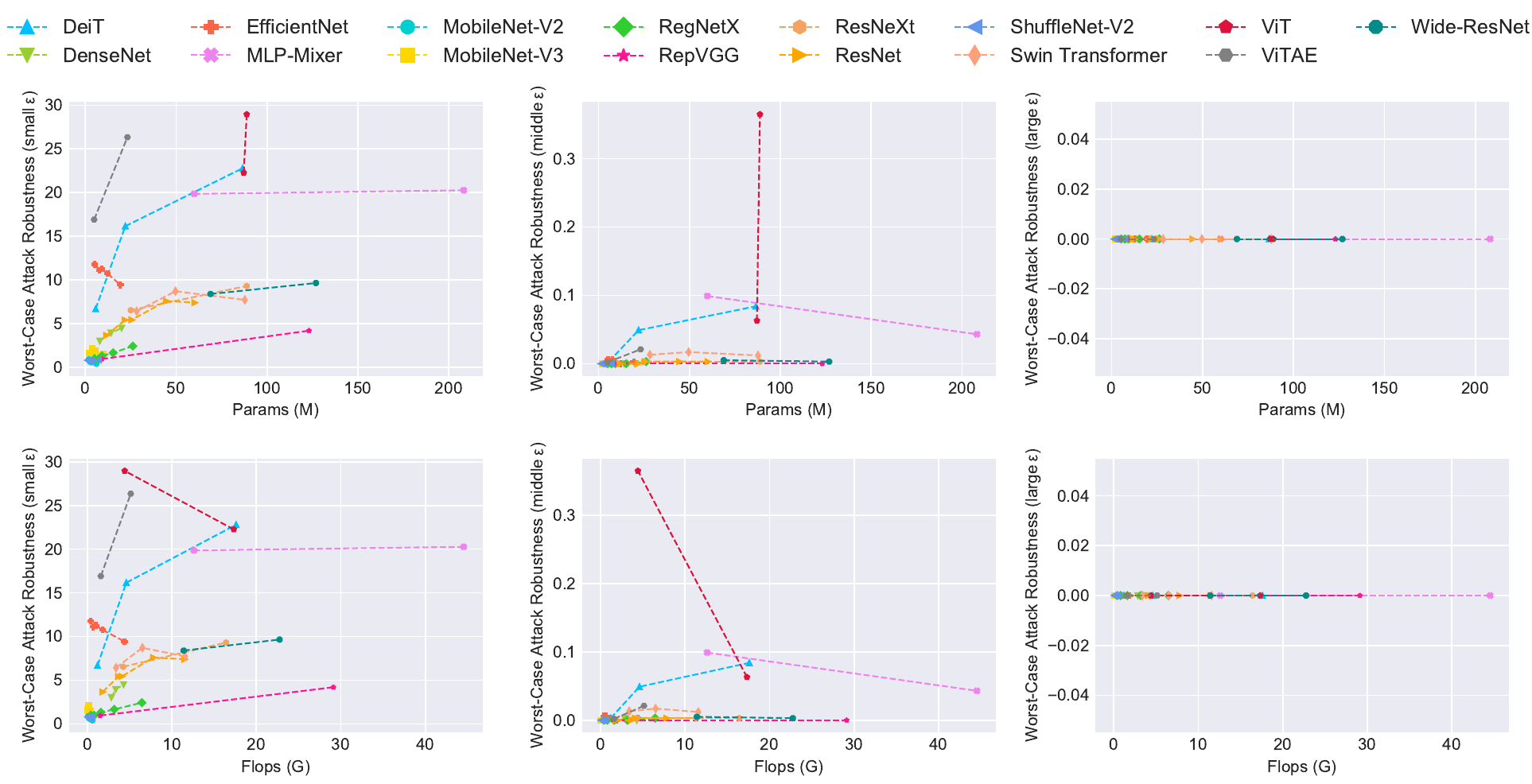}
\caption{Worst-case attack robustness of human-designed off-the-shelf architectures under 3 different $\epsilon$ magnitudes. 
}
\label{fig:supp-worstcase}
\end{figure*}

\begin{figure*}[h]
\vspace{-0.2in}
\includegraphics[width=1.0\linewidth]{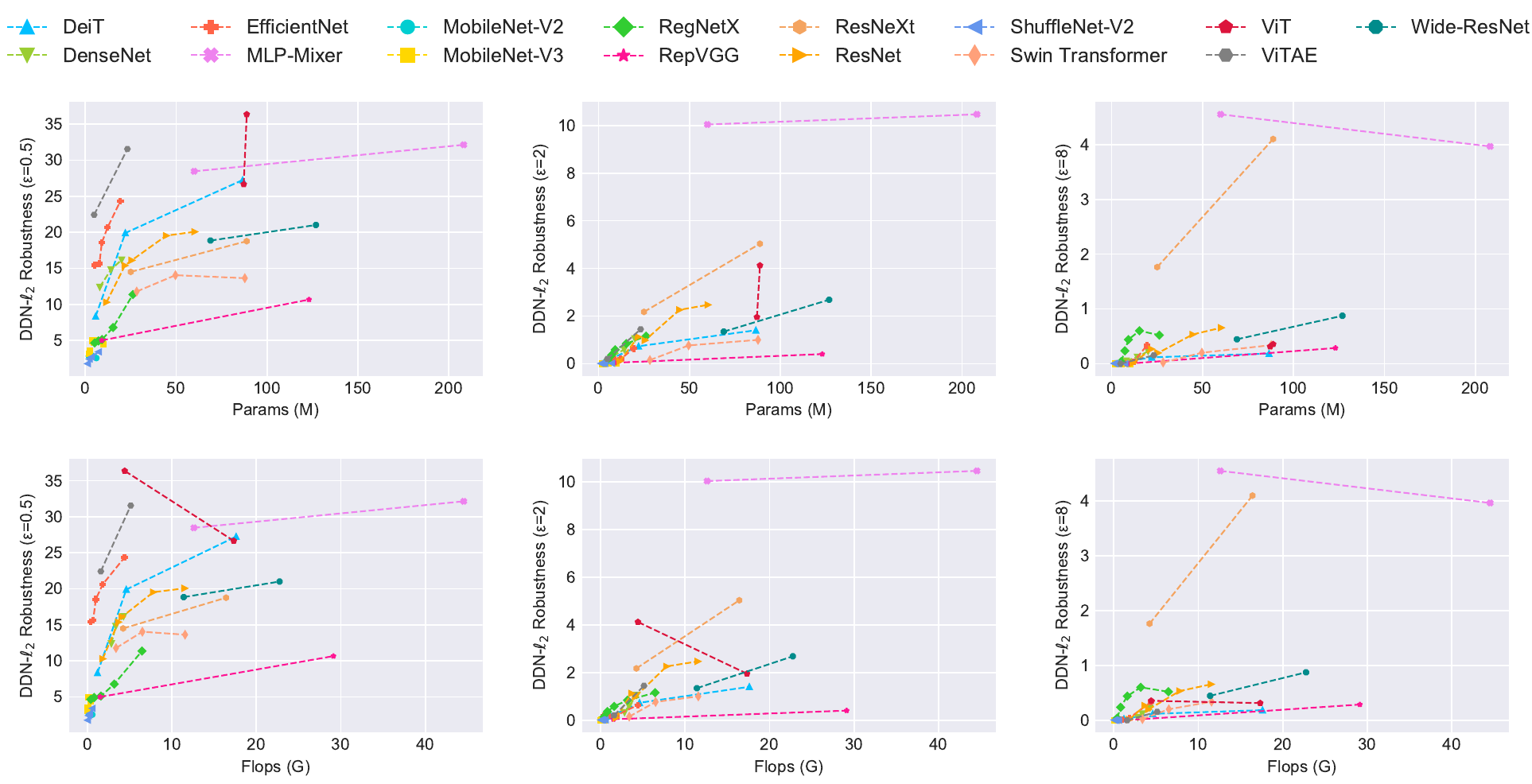}
\caption{Adversarial robustness of human-designed off-the-shelf architectures under DDN-$\ell_{2}$ attack with 3 different $\epsilon$ magnitudes. 
}
\label{fig:supp-ddn}
\end{figure*}

\begin{figure*}[h]
\vspace{-0.2in}
\includegraphics[width=1.0\linewidth]{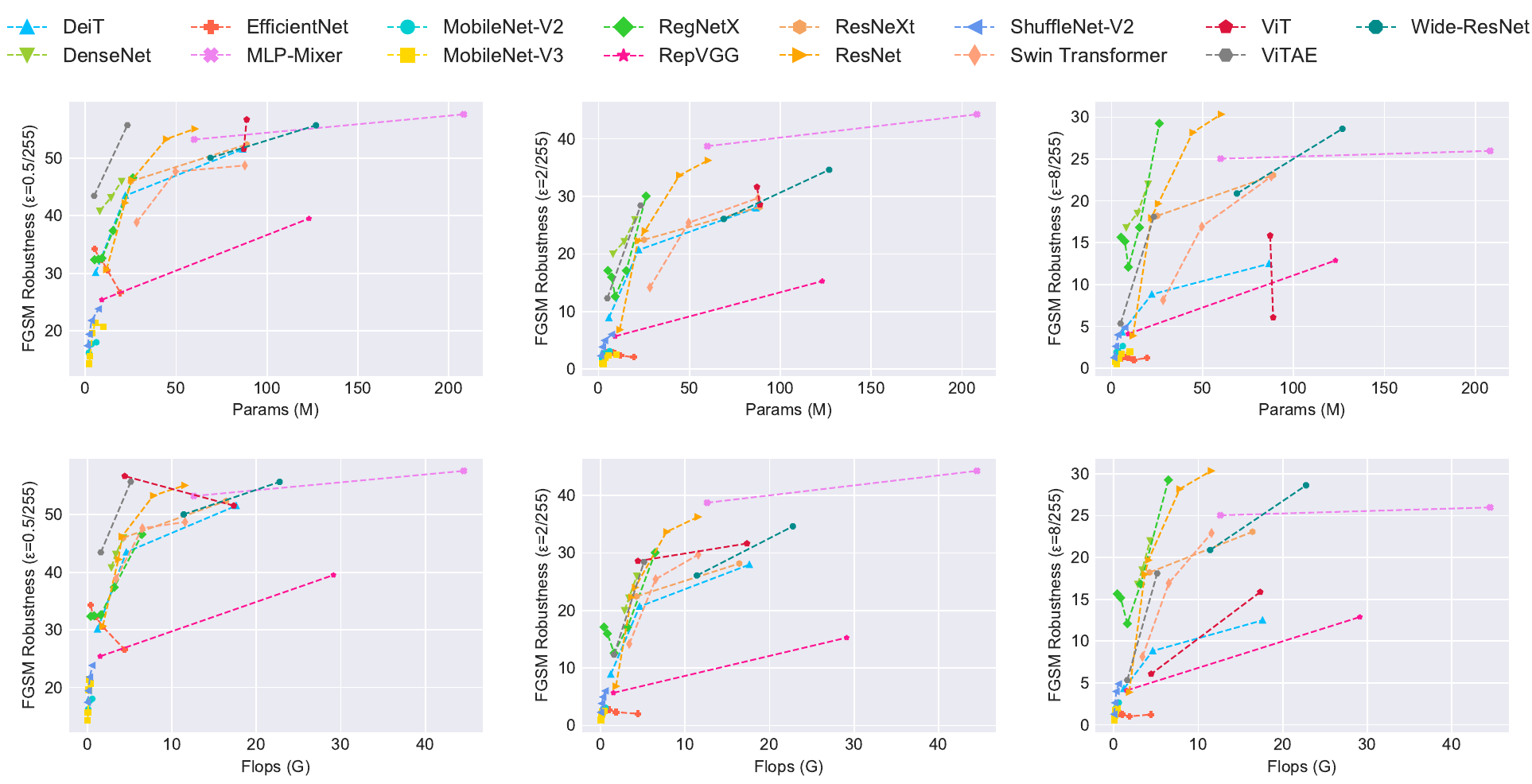}
\caption{Adversarial robustness of human-designed off-the-shelf architectures under FGSM attack with 3 different $\epsilon$ magnitudes. 
}
\label{fig:supp-fgsm}
\end{figure*}

\begin{figure*}[h]
\vspace{-0.2in}
\includegraphics[width=1.0\linewidth]{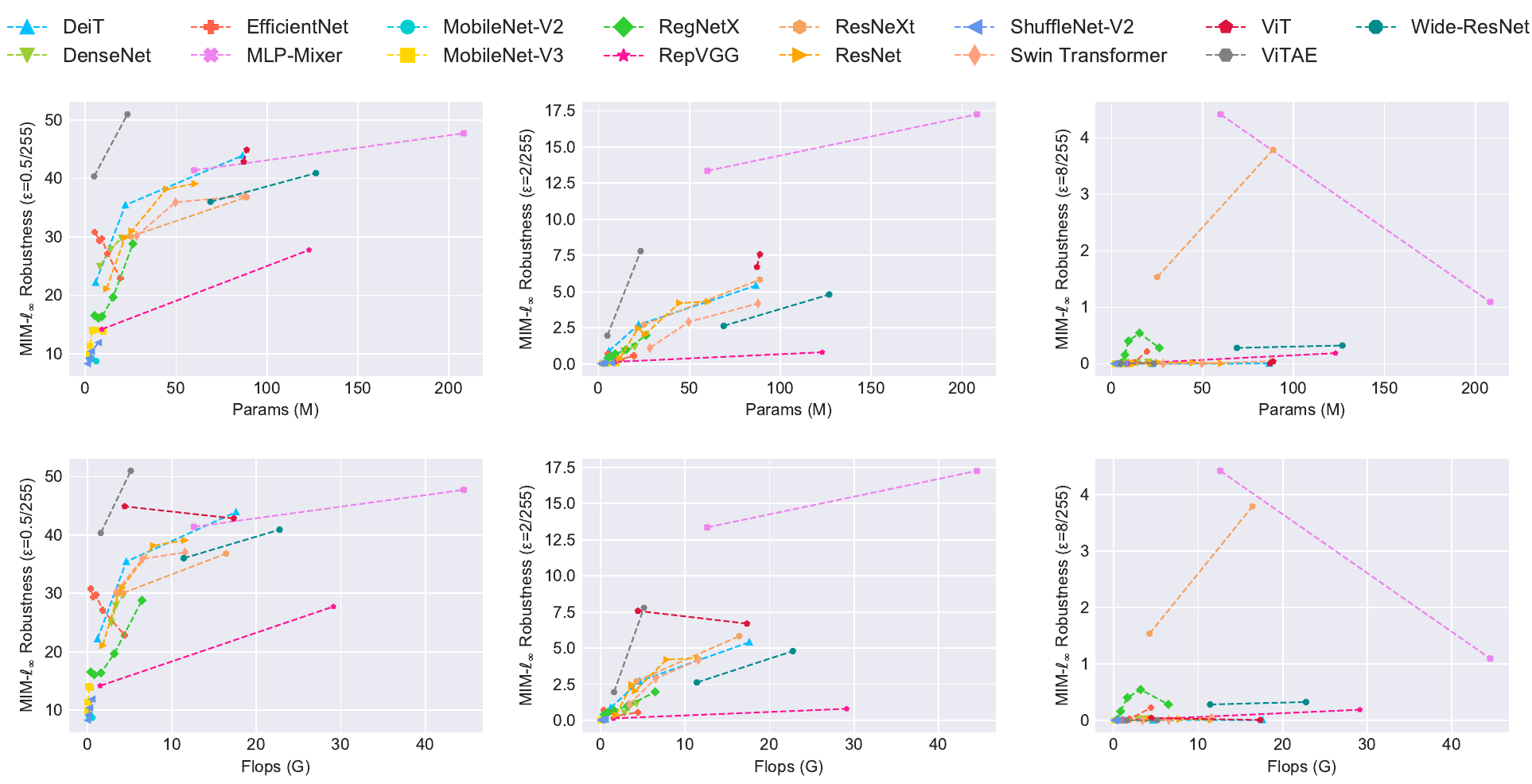}
\caption{Adversarial robustness of human-designed off-the-shelf architectures under MIM-$\ell_{\infty}$ attack with 3 different $\epsilon$ magnitudes. 
}
\label{fig:supp-mim}
\end{figure*}

\begin{figure*}[h]
\vspace{-0.2in}
\includegraphics[width=1.0\linewidth]{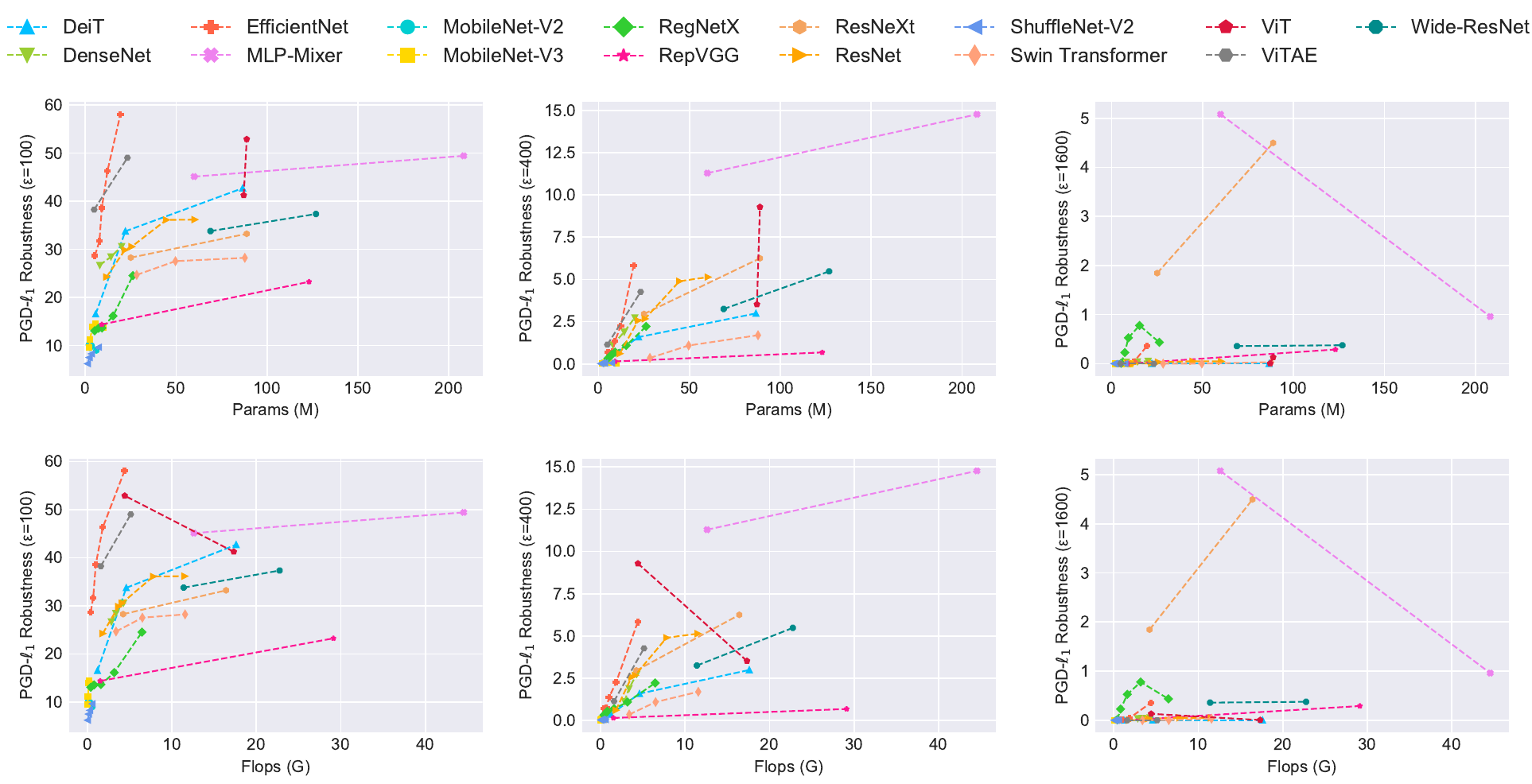}
\caption{Adversarial robustness of human-designed off-the-shelf architectures under PGD-$\ell_{1}$ attack with 3 different $\epsilon$ magnitudes. 
}
\label{fig:supp-pgdl1}
\end{figure*}

\begin{figure*}[h]
\vspace{-0.2in}
\includegraphics[width=1.0\linewidth]{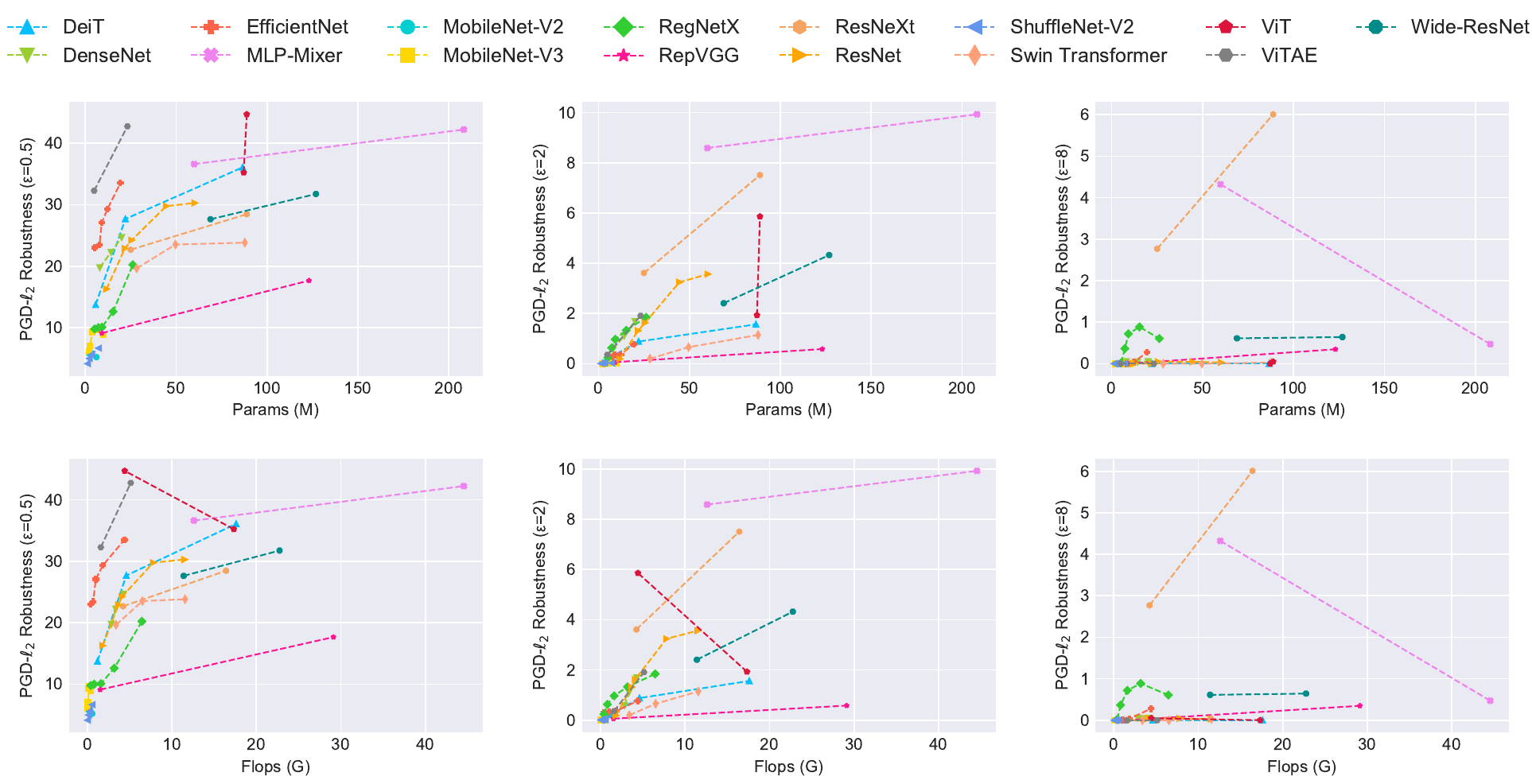}
\caption{Adversarial robustness of human-designed off-the-shelf architectures under PGD-$\ell_{2}$ attack with 3 different $\epsilon$ magnitudes. 
}
\label{fig:supp-pgdl2}
\end{figure*}

\begin{figure*}[h]
\vspace{-0.2in}
\includegraphics[width=1.0\linewidth]{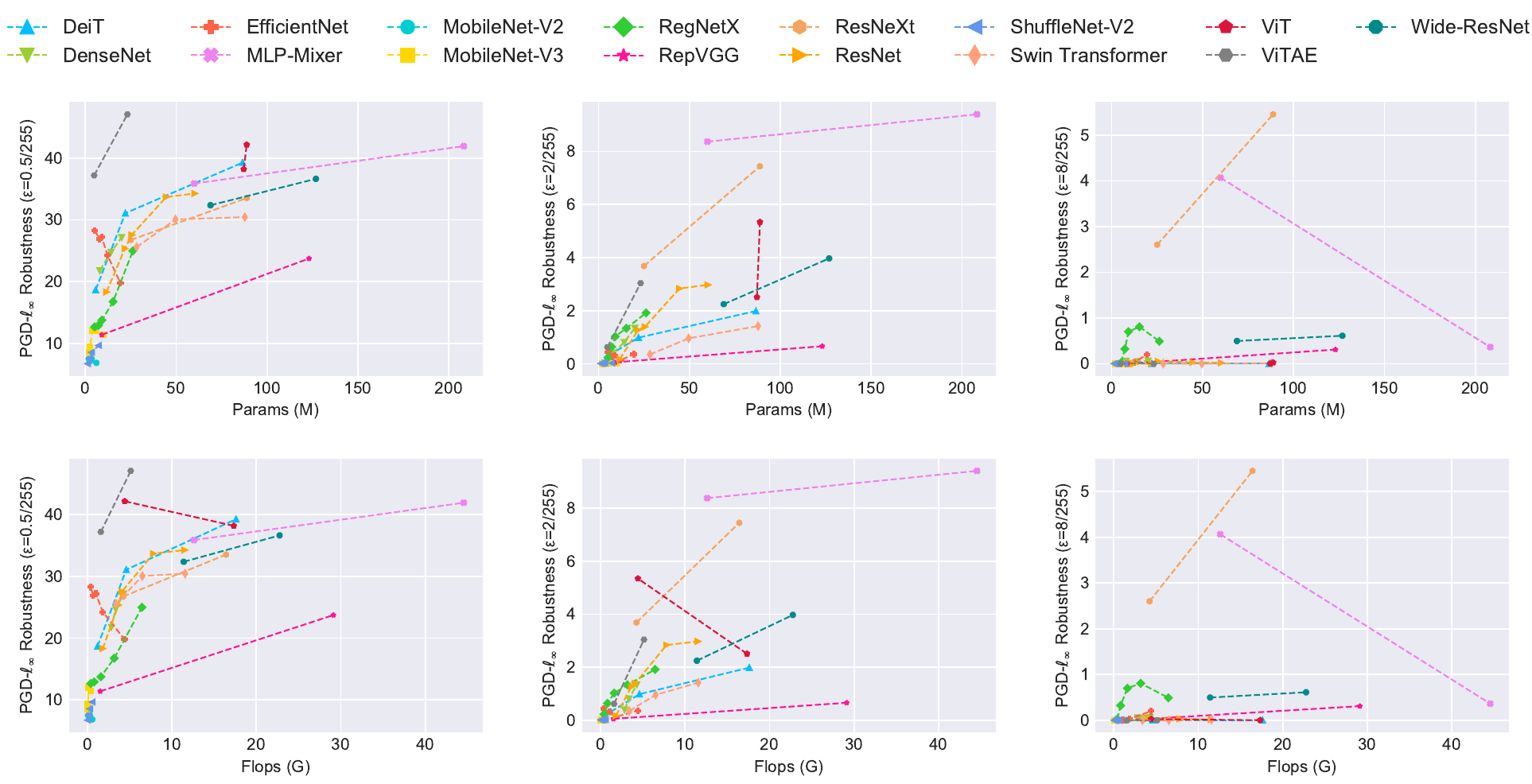}
\caption{Adversarial robustness of human-designed off-the-shelf architectures under PGD-$\ell_{\infty}$ attack with 3 different $\epsilon$ magnitudes. 
}
\label{fig:supp-pgdlinf}
\end{figure*}

\begin{figure*}[h]
\vspace{-0.2in}
\includegraphics[width=1.0\linewidth]{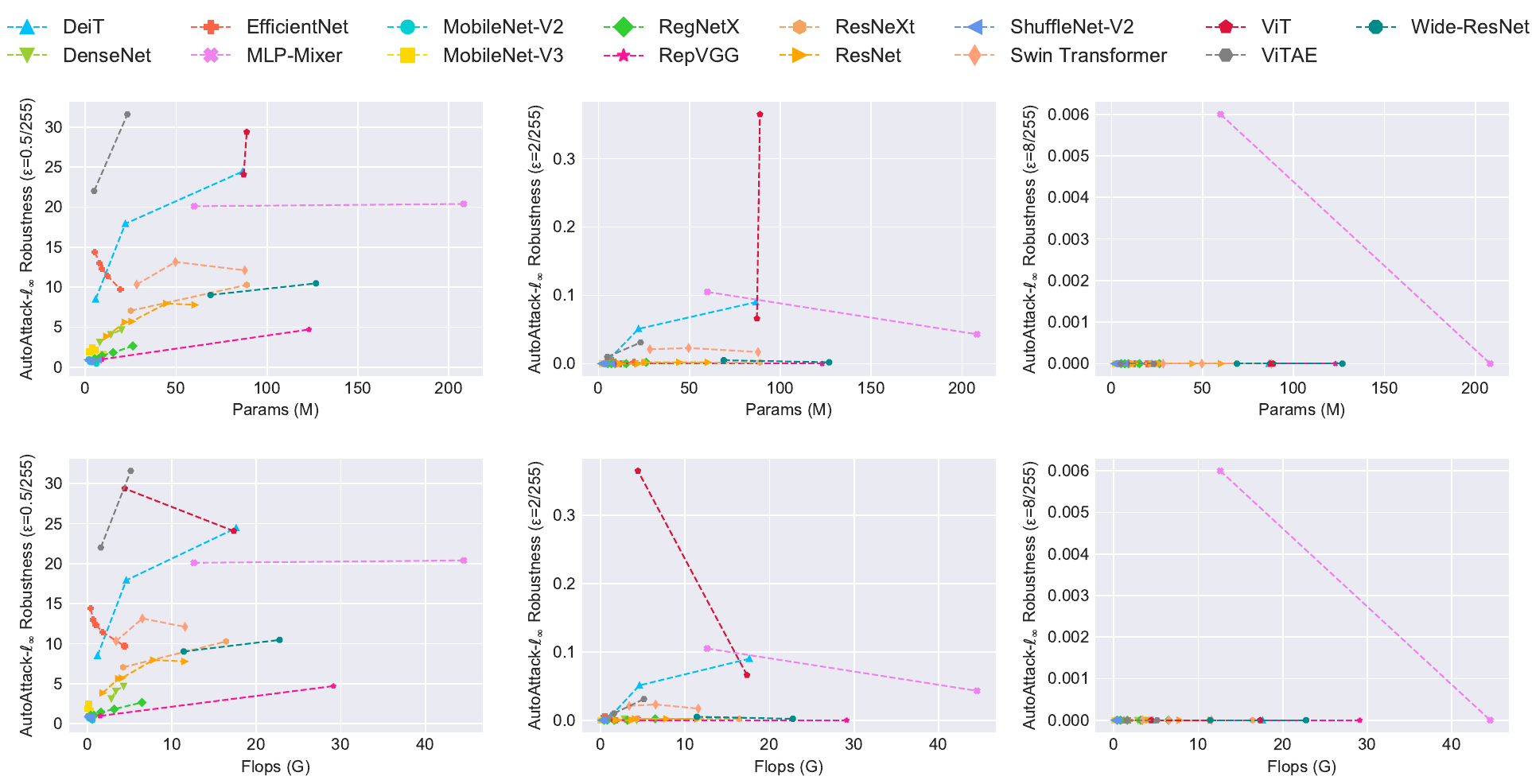}
\caption{Adversarial robustness of human-designed off-the-shelf architectures under AutoAttack-$\ell_{\infty}$ attack with 3 different $\epsilon$ magnitudes. 
}
\label{fig:supp-autoattack}
\end{figure*}

\begin{figure*}[h]
\vspace{-0.2in}
\includegraphics[width=1.0\linewidth]{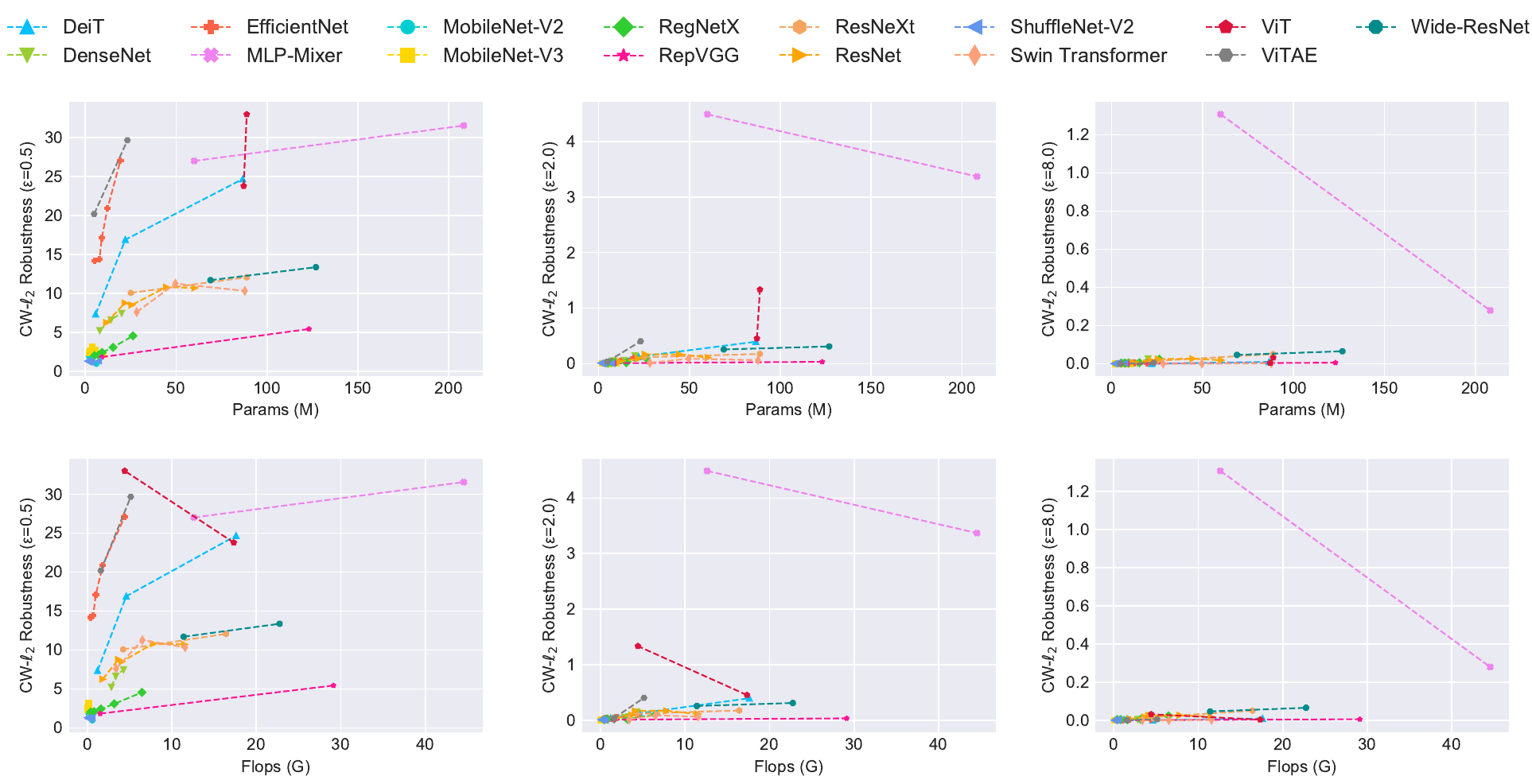}
\caption{Adversarial robustness of human-designed off-the-shelf architectures under C\&W-$\ell_{2}$ attack with 3 different $\epsilon$ magnitudes. 
}
\label{fig:supp-cw}
\end{figure*}

\begin{figure*}[h]
\vspace{-0.15in}
\includegraphics[width=1.0\linewidth]{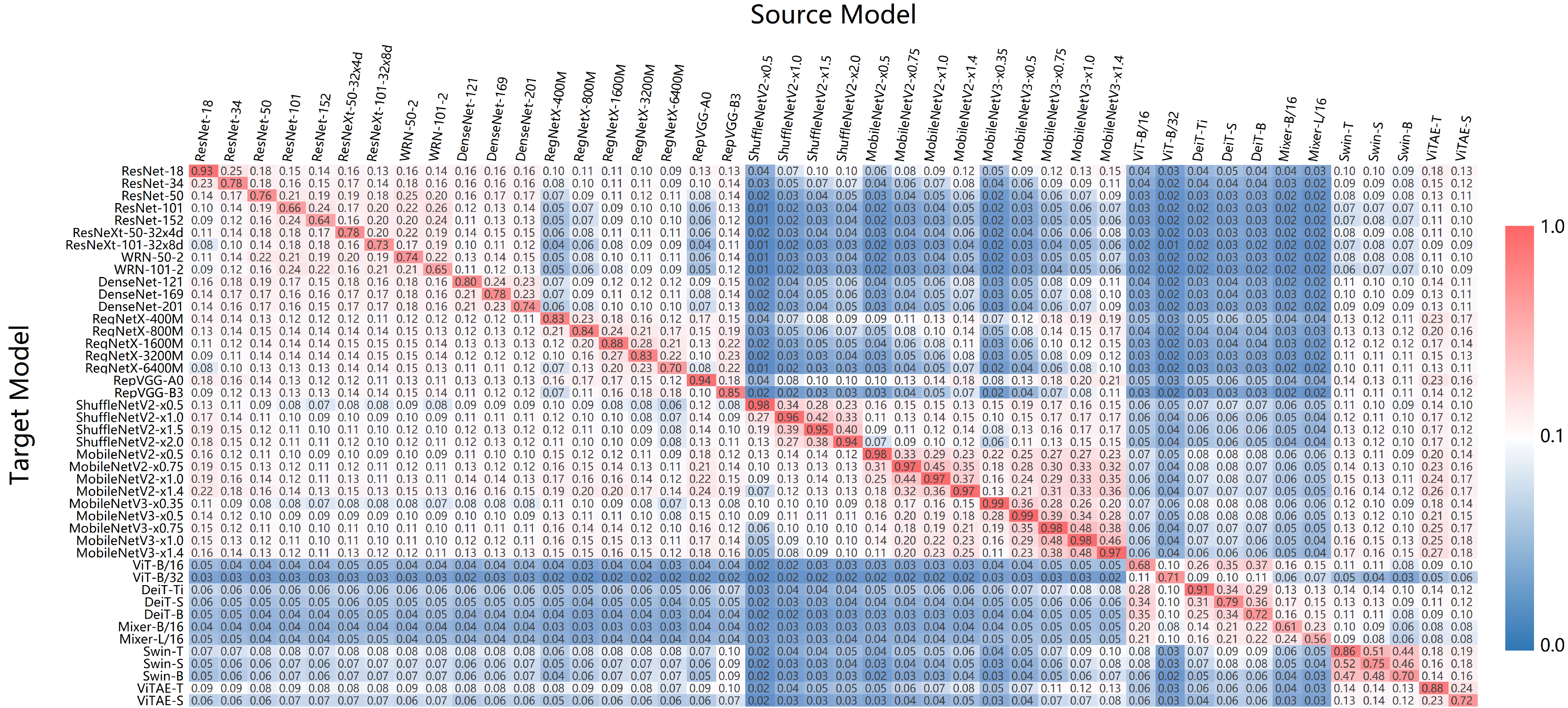}
\caption{Transferability heatmap of human-designed off-the-shelf architectures under FGSM attack, $\epsilon$=2/255.}
\label{fig:supp-transfer-fgsm2}
\vspace{-0.15in}
\end{figure*}

\begin{figure*}[h]
\vspace{-0.15in}
\includegraphics[width=1.0\linewidth]{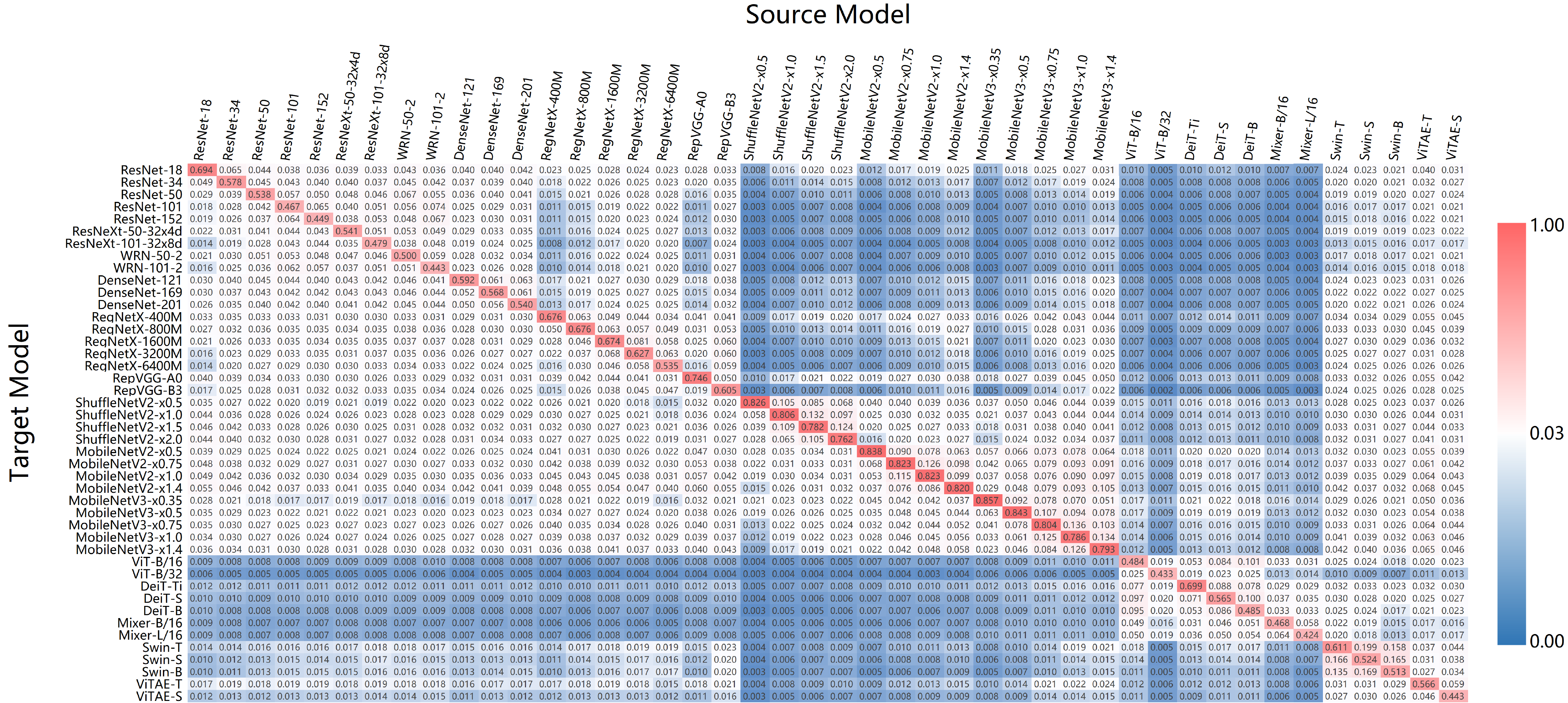}
\caption{Transferability heatmap of human-designed off-the-shelf architectures under FGSM attack, $\epsilon$=0.5/255.}
\label{fig:supp-transfer-fgsm05}
\vspace{-0.15in}
\end{figure*}

We first report the model robustness and standard performance under the view of FLOPs in Figure \ref{fig:supp-handcraft-flops} . For almost all model architectures, the robustness results are the same as those using Params to measure model size. Then we report the detailed results of model robustness under various magnitudes of different adversarial noises in Figure \ref{fig:supp-worstcase}, \ref{fig:supp-ddn}, \ref{fig:supp-fgsm}, \ref{fig:supp-mim}, \ref{fig:supp-pgdl1}, \ref{fig:supp-pgdl2}, \ref{fig:supp-pgdlinf}, \ref{fig:supp-autoattack}, \ref{fig:supp-cw}. There are 2 model size measurements (FLOPs and Params) and 3 perturbation magnitudes (small, middle and large), resulting in total 6 figure for each attack. In addition, we also report the heatmap of different model architectures under transfer-based adversarial attacks under more perturbation magnitudes in Figure \ref{fig:supp-transfer-fgsm2}, \ref{fig:supp-transfer-fgsm05}. We can see the model robustness results under transfer-based attack with $\epsilon$=0.5/255 and $\epsilon$=2/255 are almost the same as those under attack with $\epsilon$=8/255. 

\subsection{Training Techniques Towards Robustness}
\label{sec:supp_technique-moreres}
\begin{figure*}[h]
\vspace{-0.2in}
\includegraphics[width=1.0\linewidth]{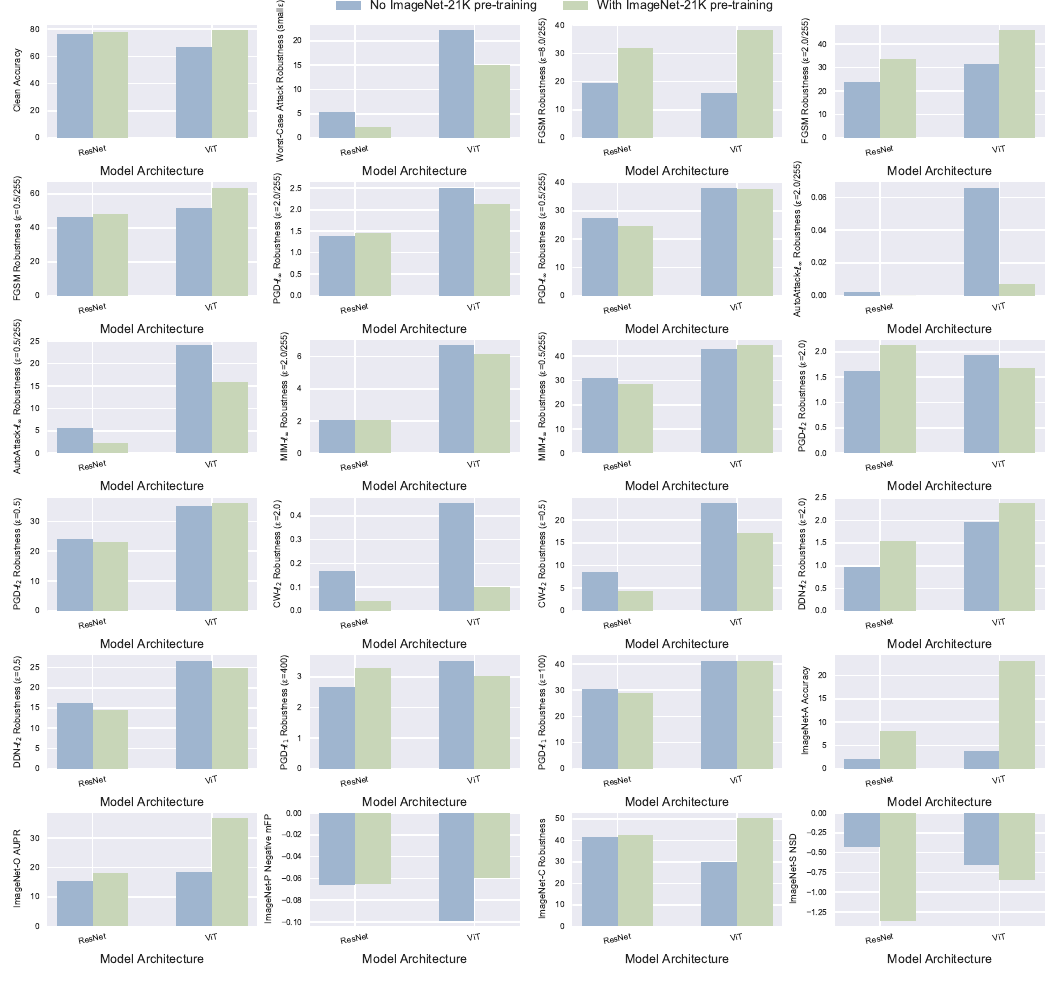}
\caption{The influence of ImageNet-21K pre-training towards model robustness under various noises. 
}
\label{fig:supp-21kpretrain}
\end{figure*}

\begin{figure*}[h]
\vspace{-0.2in}
\includegraphics[width=1.0\linewidth]{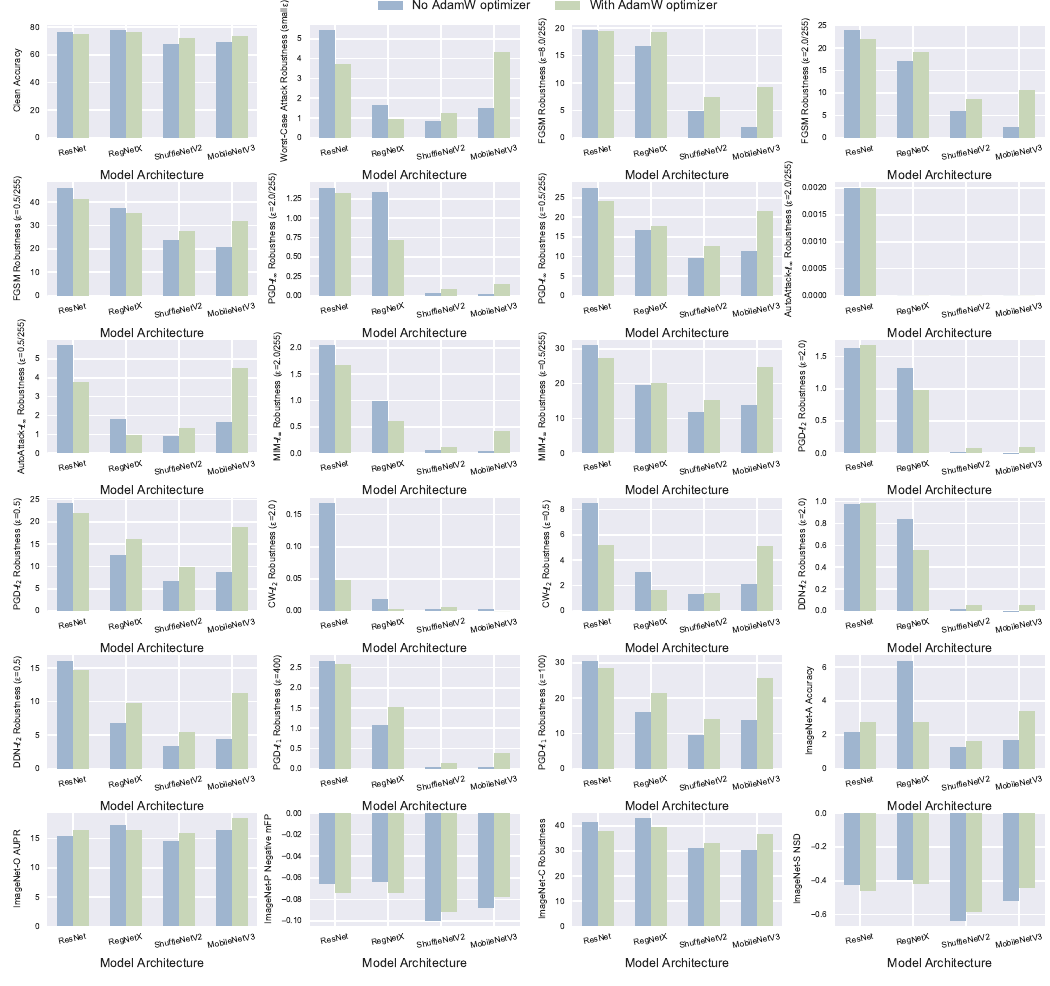}
\caption{The influence of AdamW optimizer towards model robustness under various noises. 
}
\label{fig:supp-adamw}
\end{figure*}

\begin{figure*}[h]
\vspace{-0.2in}
\includegraphics[width=1.0\linewidth]{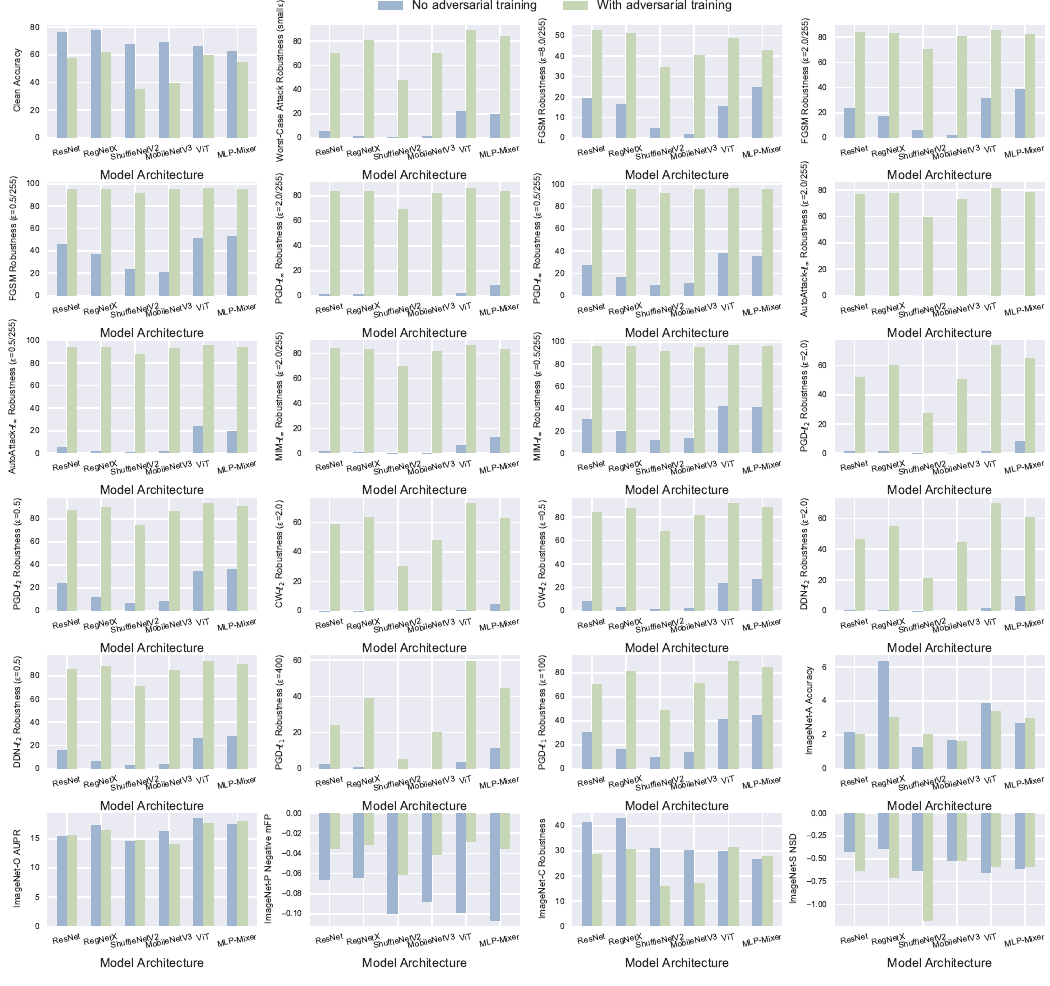}
\caption{The influence of adversarial training towards model robustness under various noises. 
}
\label{fig:supp-advtrain}
\end{figure*}

\begin{figure*}[h]
\vspace{-0.2in}
\includegraphics[width=1.0\linewidth]{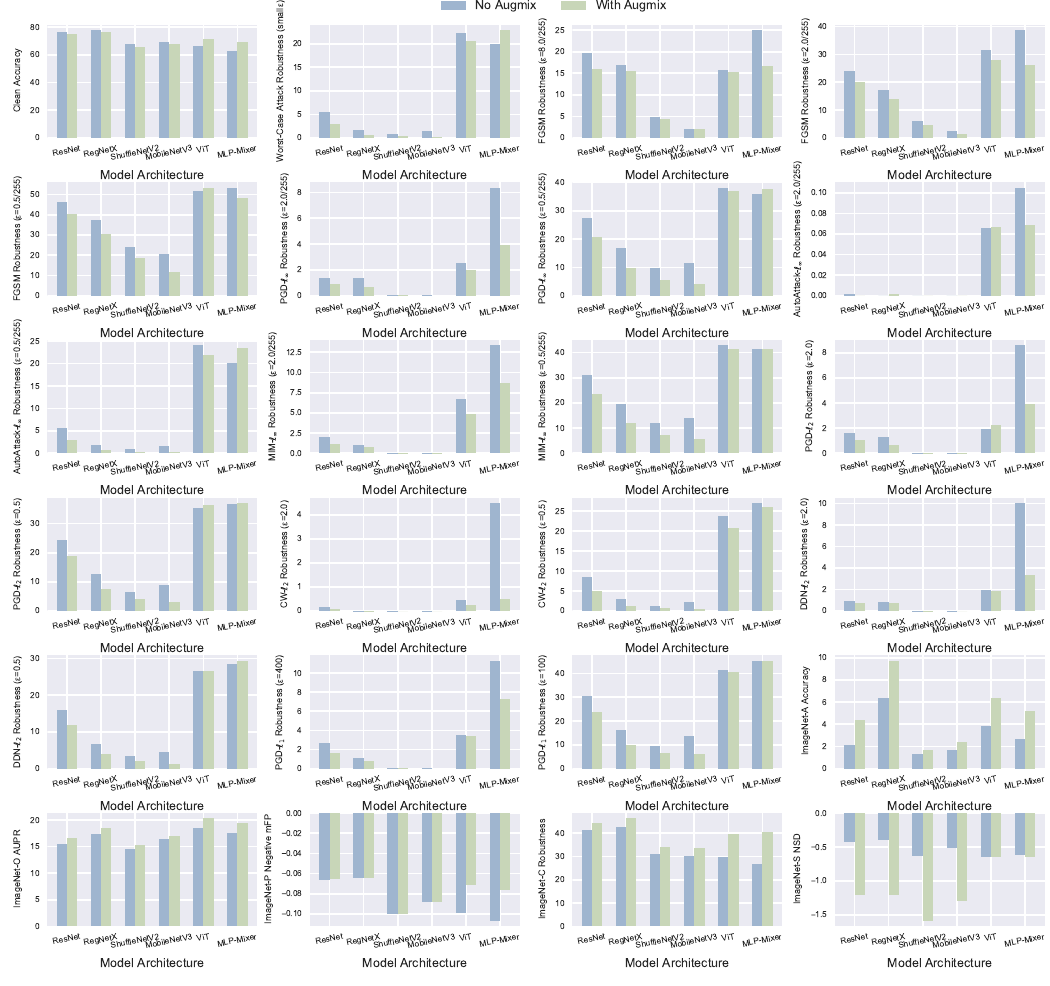}
\caption{The influence of Augmix data augmentation towards model robustness under various noises. 
}
\label{fig:supp-augmix}
\end{figure*}

\begin{figure*}[h]
\vspace{-0.2in}
\includegraphics[width=1.0\linewidth]{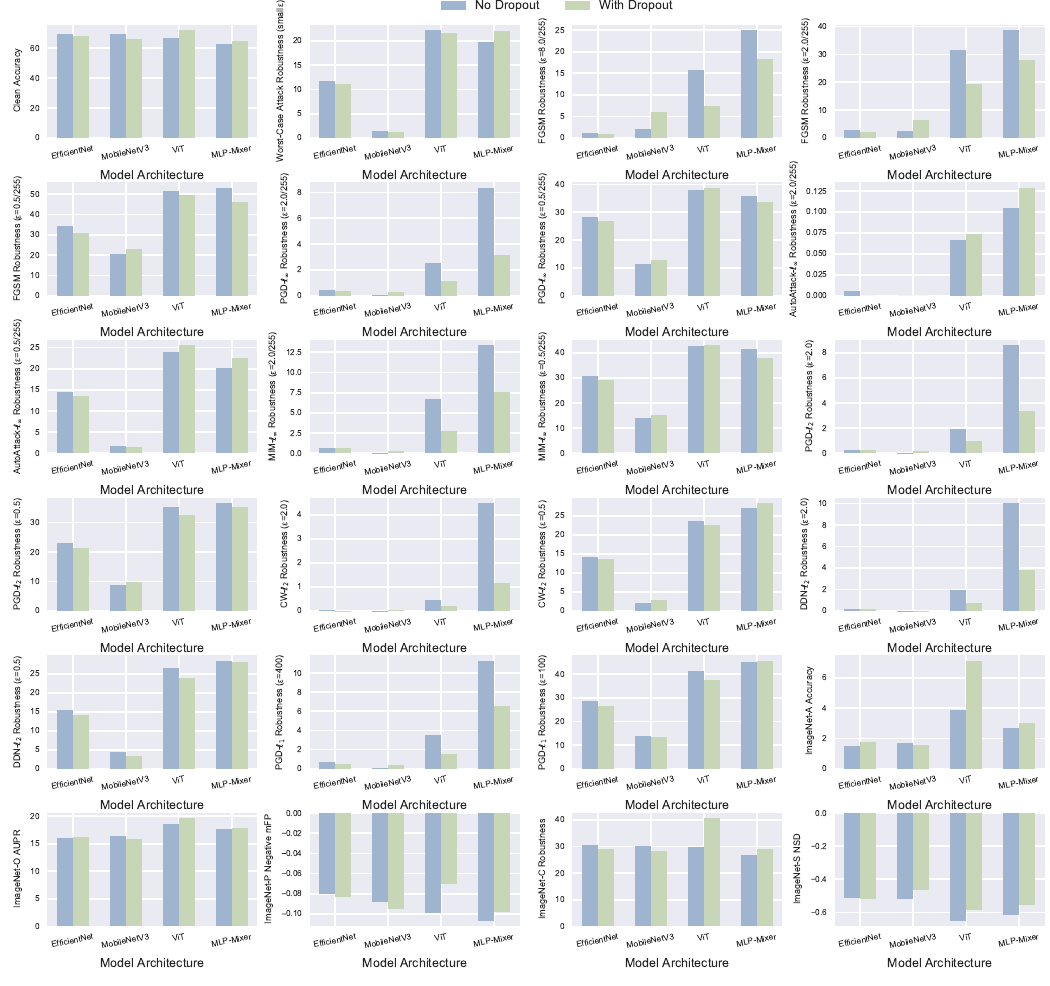}
\caption{The influence of Dropout towards model robustness under various noises. 
}
\label{fig:supp-dropout}
\end{figure*}

\begin{figure*}[h]
\vspace{-0.2in}
\includegraphics[width=1.0\linewidth]{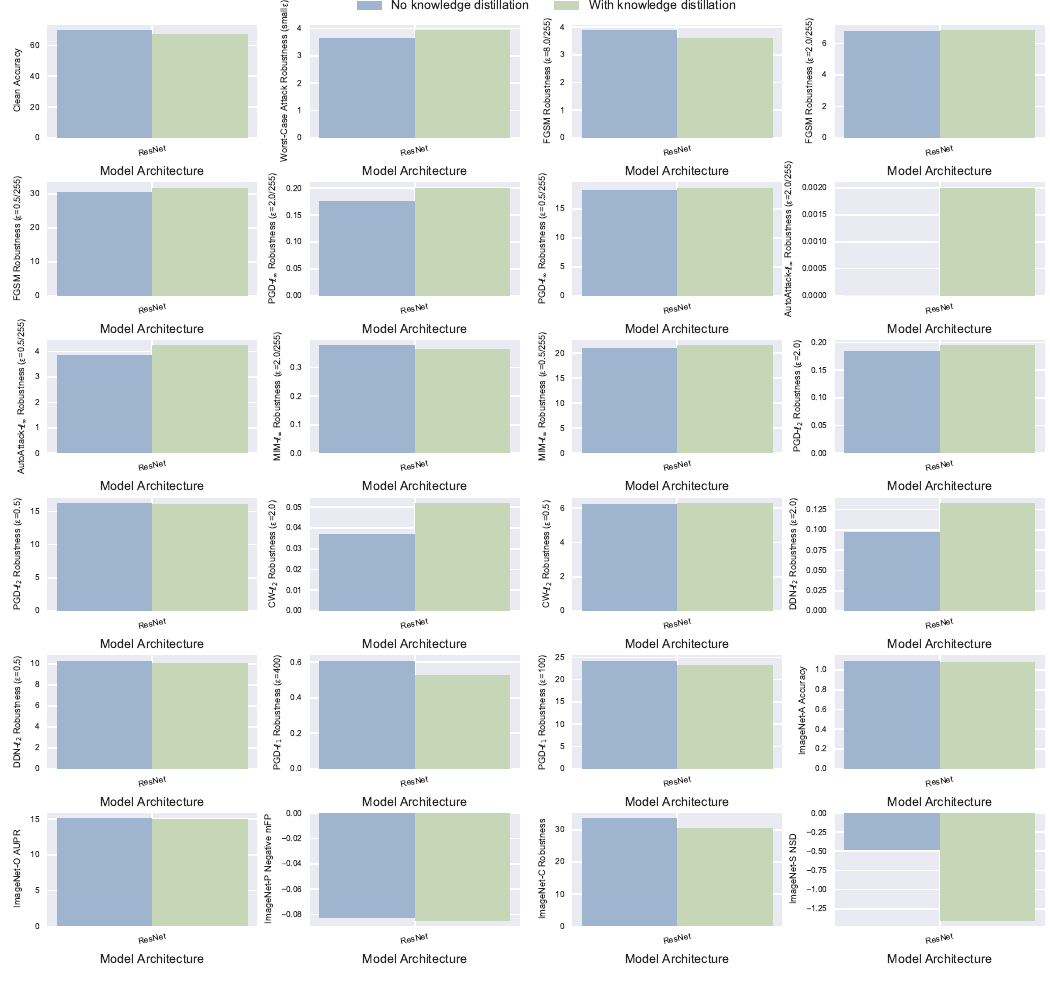}
\caption{The influence of knowledge distillation towards model robustness under various noises. 
}
\label{fig:supp-kd}
\end{figure*}

\begin{figure*}[h]
\vspace{-0.2in}
\includegraphics[width=1.0\linewidth]{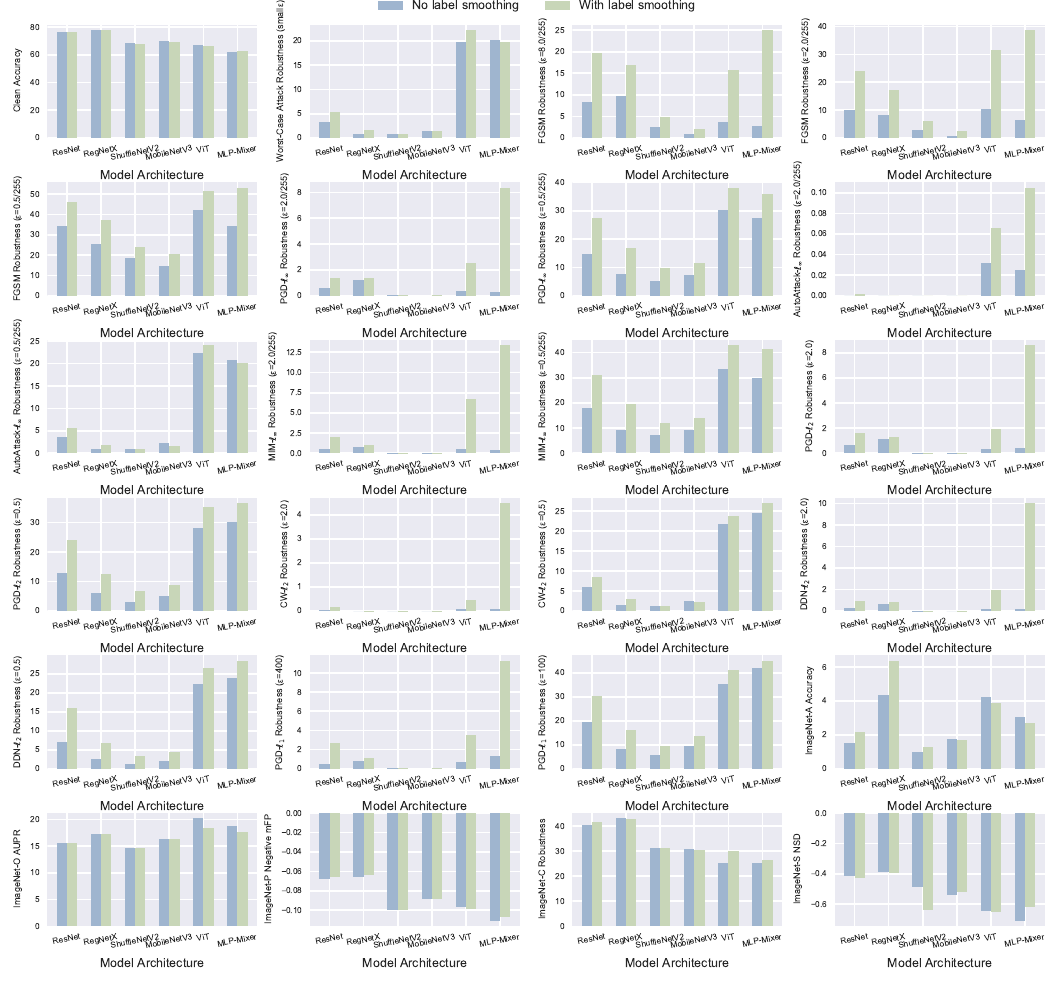}
\caption{The influence of label smoothing towards model robustness under various noises. 
}
\label{fig:supp-labelsmooth}
\end{figure*}

\begin{figure*}[h]
\vspace{-0.2in}
\includegraphics[width=1.0\linewidth]{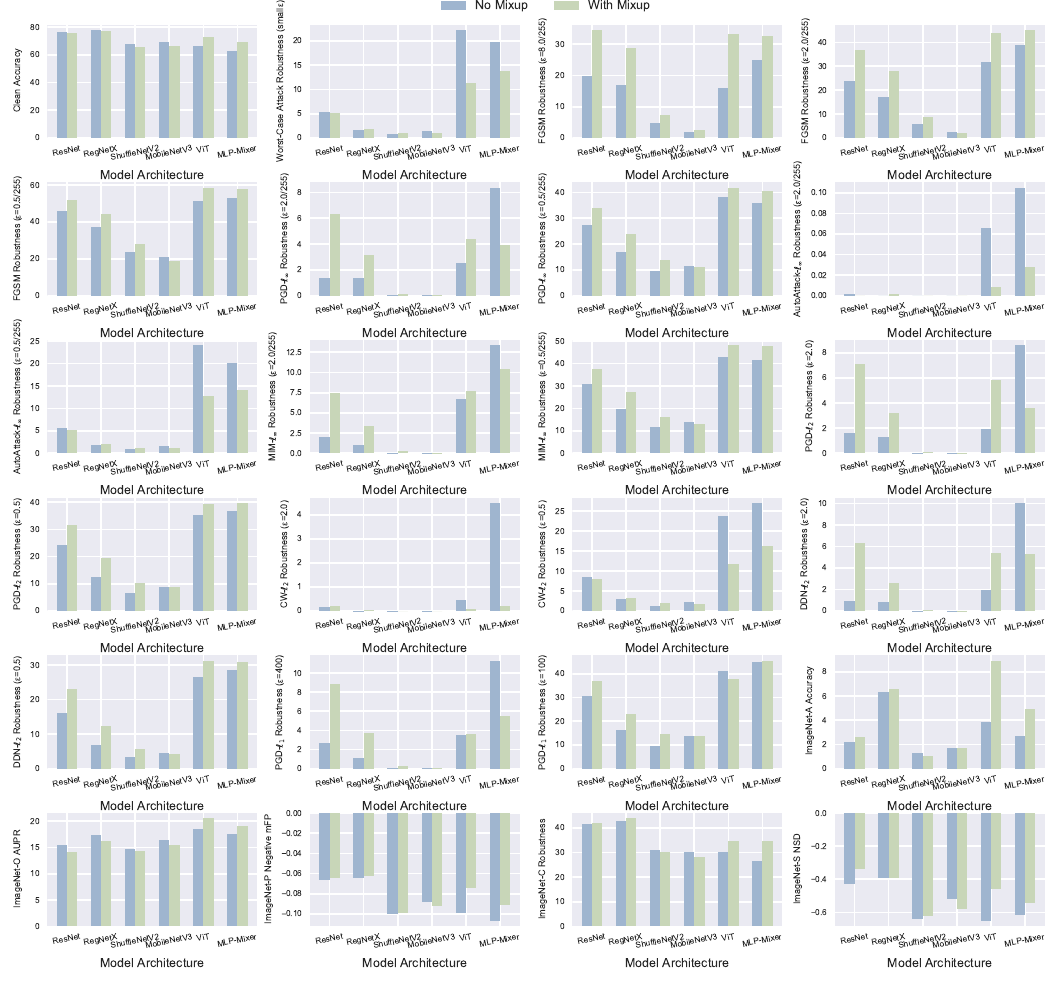}
\caption{The influence of Mixup data augmentation towards model robustness under various noises. 
}
\label{fig:supp-mixup}
\end{figure*}

\begin{figure*}[h]
\vspace{-0.2in}
\includegraphics[width=1.0\linewidth]{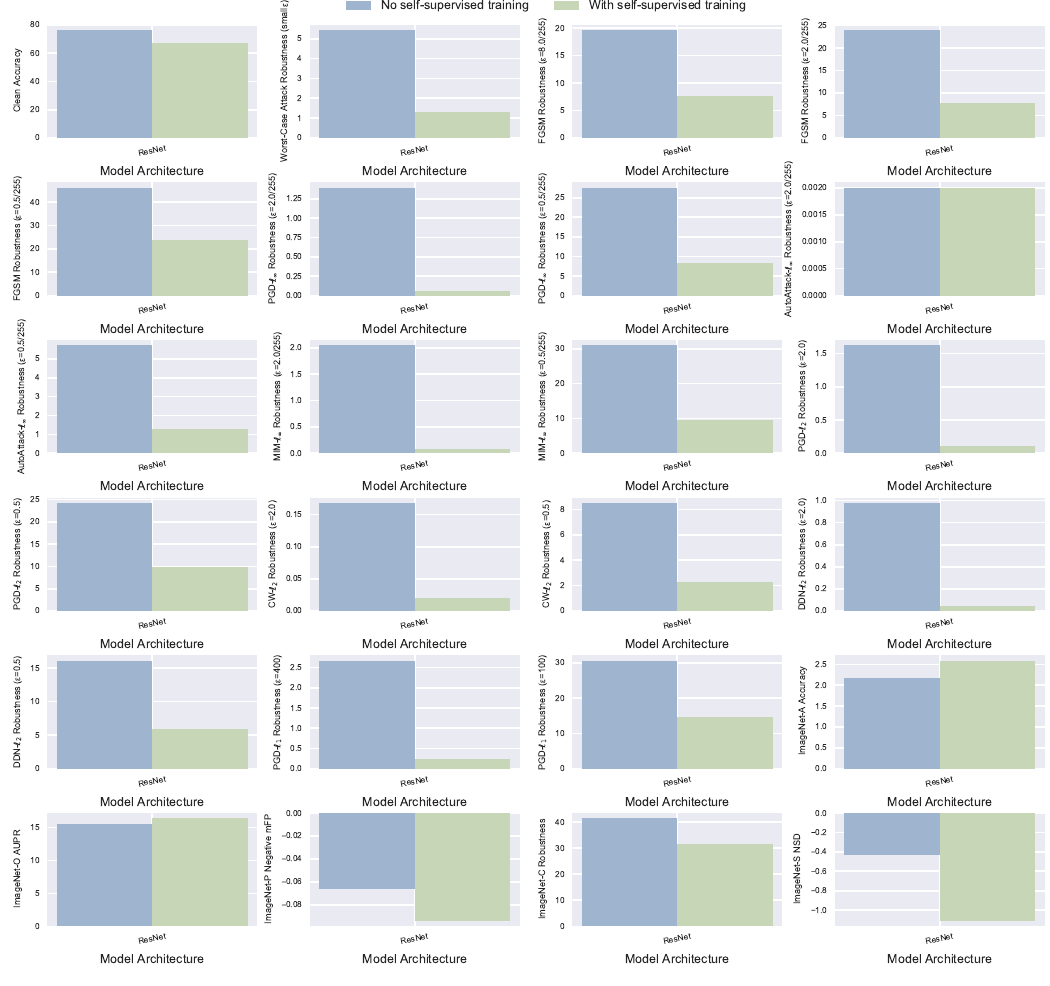}
\caption{The influence of self-supervised training towards model robustness under various noises. 
}
\label{fig:supp-selfsup}
\end{figure*}

\begin{figure*}[h]
\vspace{-0.2in}
\includegraphics[width=1.0\linewidth]{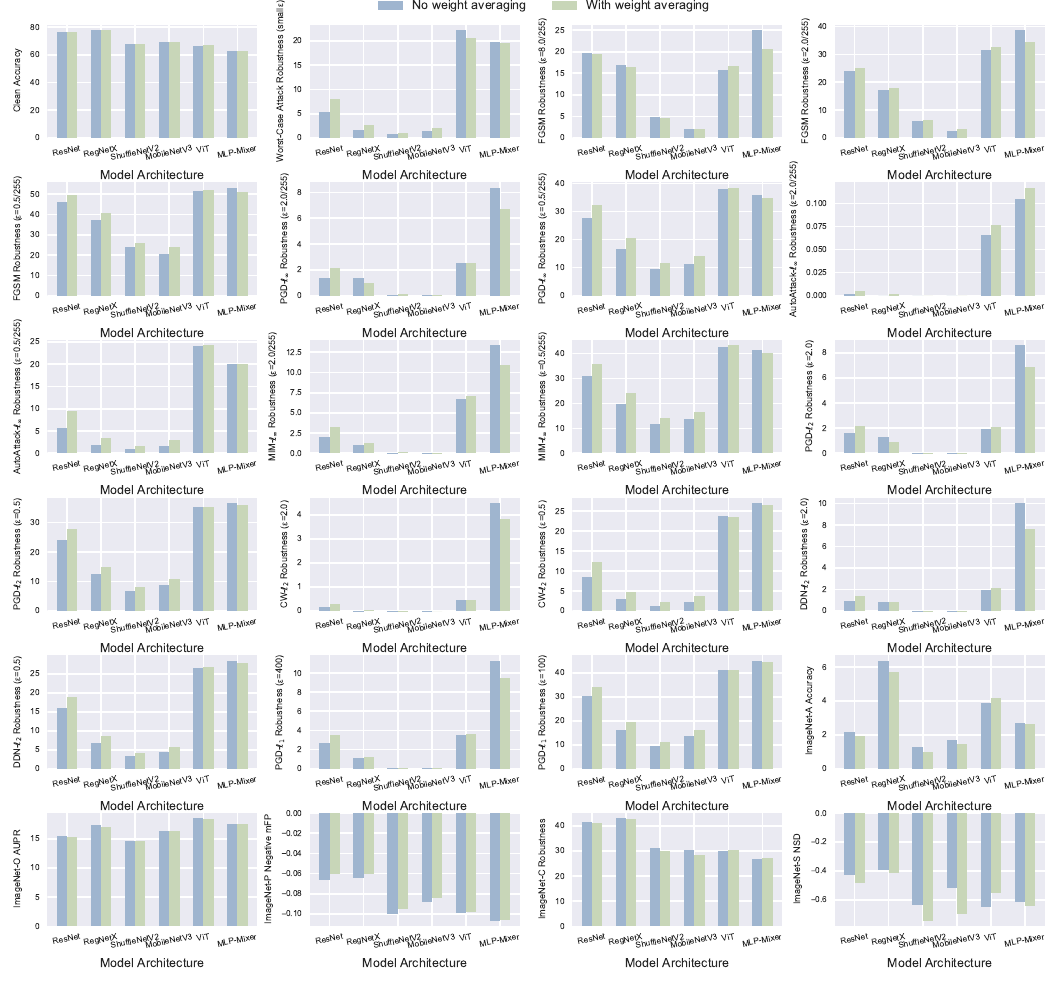}
\caption{The influence of weight averaging towards model robustness under various noises. 
}
\label{fig:supp-weightaver}
\end{figure*}

\begin{figure*}[h]
\vspace{-0.2in}
\includegraphics[width=1.0\linewidth]{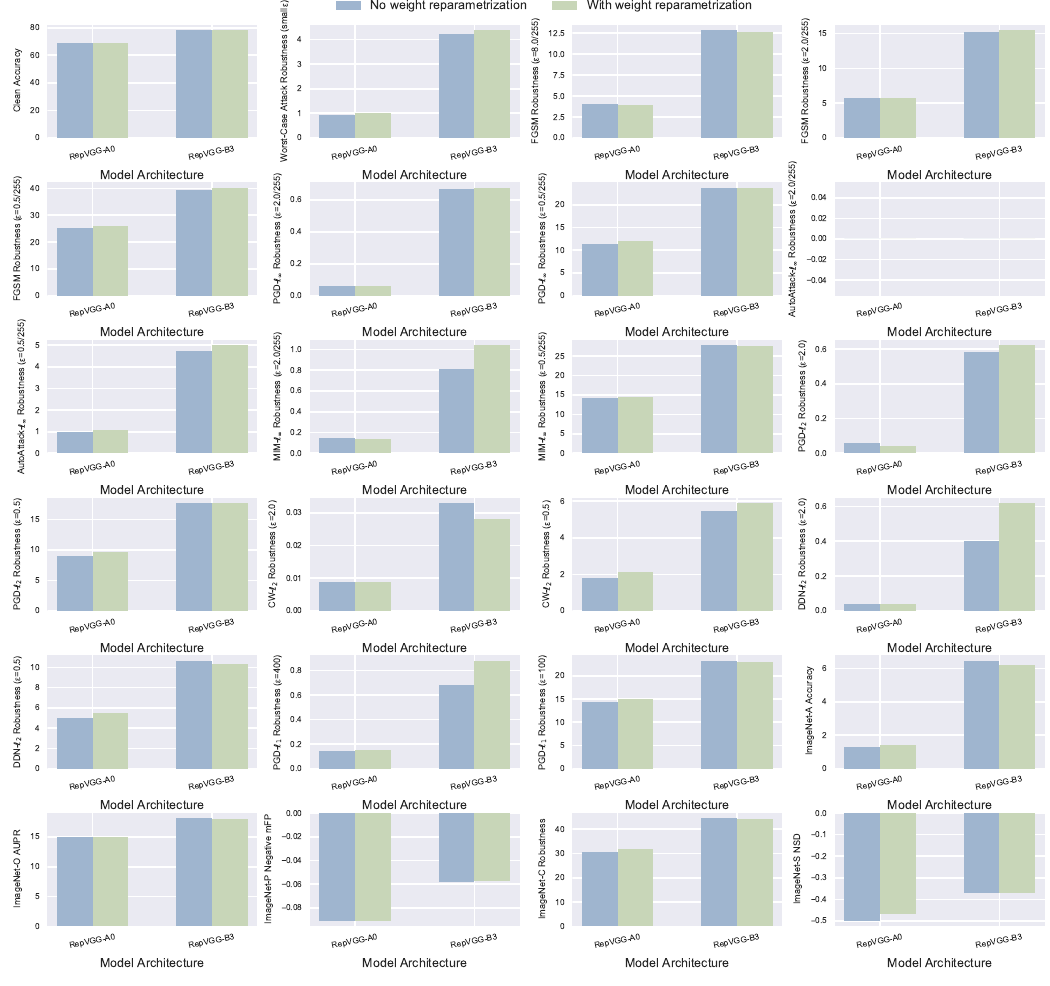}
\caption{The influence of weight re-parameterization towards model robustness under various noises. 
}
\label{fig:supp-weightrepara}
\end{figure*}

We report the results for the influence of all training techniques studied towards model robustness under various noises in Figure \ref{fig:supp-21kpretrain}, \ref{fig:supp-adamw}, \ref{fig:supp-advtrain}, \ref{fig:supp-augmix}, \ref{fig:supp-dropout}, \ref{fig:supp-kd}, \ref{fig:supp-labelsmooth}, \ref{fig:supp-mixup}, \ref{fig:supp-selfsup}, \ref{fig:supp-weightaver}, \ref{fig:supp-weightrepara}. For each training technique we choose 24 metrics to show its influence on model accuracy and robustness. More results can be found on our website.

\begin{figure*}[h]
\vspace{-0.2in}
\includegraphics[width=1.0\linewidth]{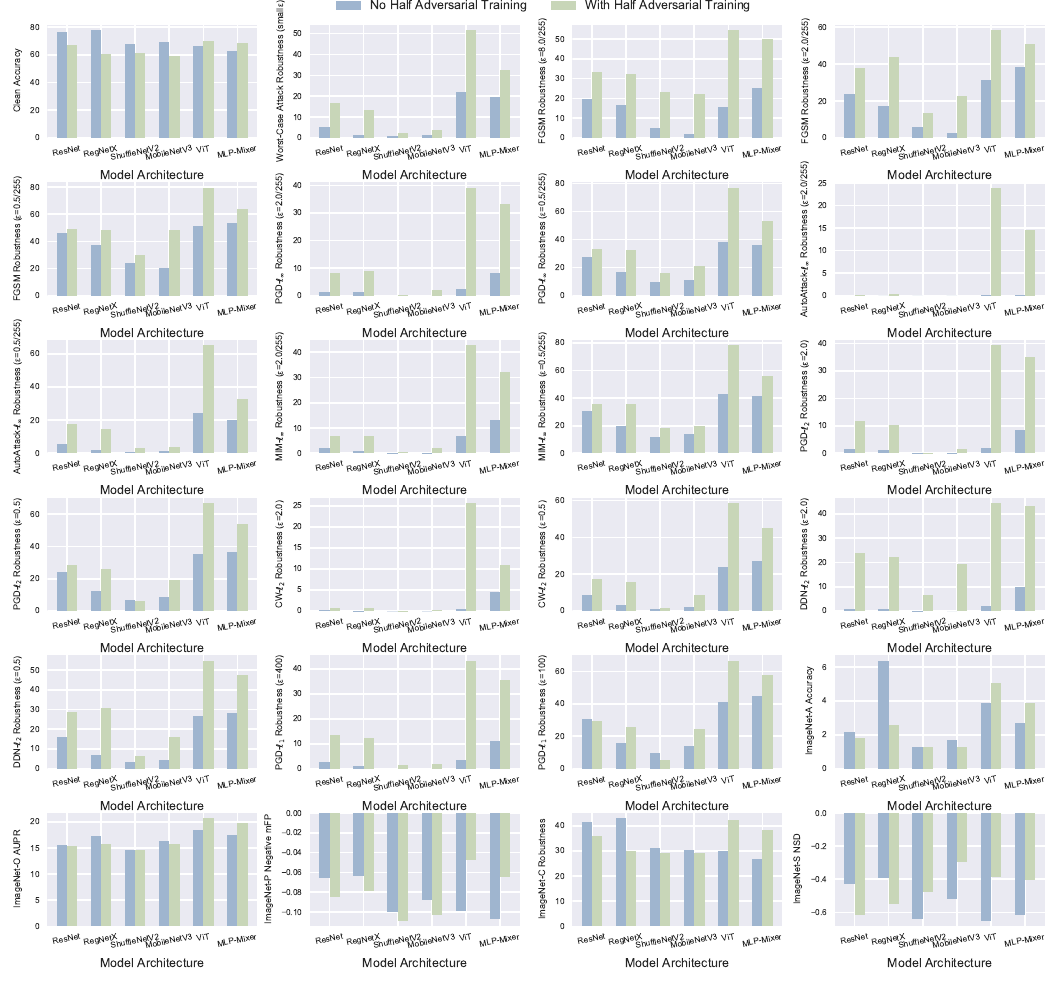}
\caption{The influence of Half Adversarial Training scheme towards model robustness under various noises. 
}
\label{fig:supp-halfadvtrain}
\end{figure*}